\pdfoutput=1
\documentclass[11pt]{article}

\usepackage[preprint]{acl}

\usepackage{times}
\usepackage{latexsym}
\usepackage[T1]{fontenc}
\usepackage[utf8]{inputenc}
\usepackage{microtype}
\usepackage{inconsolata}
\usepackage{graphicx}
\usepackage{booktabs}
\usepackage{makecell}
\usepackage{array}
\usepackage{amsmath}
\usepackage{amssymb}
\usepackage{xcolor}

\graphicspath{{./}{../../reports/figures/summary/}{../../reports/figures/}}

\newcommand{\name}{\textsc{AgentIdeaBench}}
\newcommand{\swm}{SWM}
\newcommand{\cmark}{\textcolor{green!55!black}{\checkmark}}
\newcommand{\xmark}{\textcolor{red!70!black}{$\times$}}
\newcommand{\pmark}{\textcolor{orange!85!black}{$\sim$}}

\title{\textsc{AgentIdeaBench}: Benchmarking Scientific Ideation in the Agent Era}

\author{Yunxiang Mo$^1$\thanks{~~Equal Contribution}, Tianshi Zheng$^1$\footnotemark[1], Yisen Gao$^1$, Rui Wang$^1$, Newt Nguyen Kim Hue Nam$^1$\\ \textbf{Kelvin Kiu Wai Tam$^1$, Jiaxin Bai$^1$, Yangqiu Song$^1$, Ginny Wong$^2$, Simon See$^2$} \\
  $^1$Department of Computer Science and Engineering, HKUST, Hong Kong SAR, China\\
  $^2$NVIDIA AI Technology Center (NVAITC), NVIDIA, Santa Clara, USA\\
  \texttt{\{ymoaj, tzhengad\}@connect.ust.hk, yqsong@cse.ust.hk}\\
}

\begin{document}
\maketitle
\begingroup\renewcommand{\thefootnote}{}%
\footnotetext{Code, prompts, and released data: \url{https://github.com/HKUST-KnowComp/AgentIdeaBench}}%
\endgroup

\begin{abstract}
Scientific ideation is the capacity to formulate novel and testable hypotheses from scientific evidence, and autonomous AI scientists depend on it. Existing evaluations largely assess it by asking models to generate ideas from a static, curated set of reference papers. That passive setup departs from the retrieval-and-reasoning workflow of modern AI scientists, and it becomes less discriminative as models improve. We introduce \name{}, a multidisciplinary benchmark that evaluates scientific ideation under two matched settings, \textbf{static observation} and \textbf{active exploration}. We report matched Static--Active evaluations for 33 LLMs across 40 densely scored subfields spanning five disciplines, using a multidimensional, literature-verified scoring framework whose critics assess originality against retrieved prior art. Active exploration reveals considerably more capability headroom, and that headroom is unevenly distributed across models. Performance scales about twice as fast as under static observation, and the exploration gain is capability-gated, favoring the strongest models over the weakest. The gain reflects better grounding, improving feasibility, clarity, and specificity while leaving measured originality unchanged under our critics. We further explore Scientific World Modeling, a generation-time loop that refines a draft hypothesis through structured thought experiments. It benefits mid-capability models, and its impact diminishes among frontier models that appear to have internalized such reasoning patterns already. \name{} gives future work on scientific ideation a measurement basis suited to the agent era.
\end{abstract}

\section{Introduction}\label{sec:intro}


\begin{figure}[t!]
\centering
\includegraphics[width=\columnwidth]{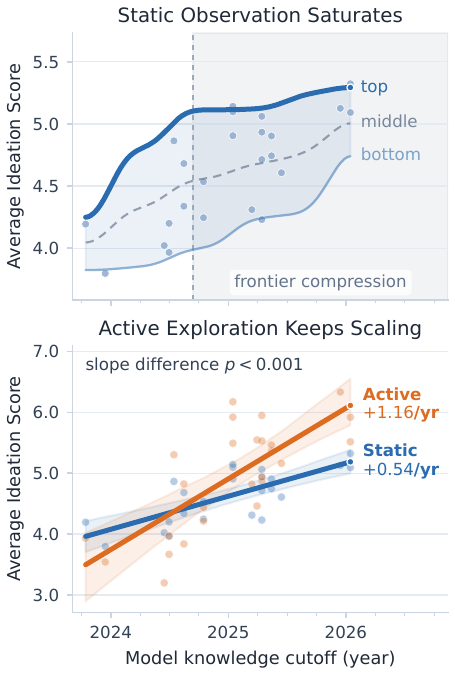}
\caption{Average ideation score against model knowledge cutoff, one point per model, over the primary roster of Table~\ref{tab:models}. \textbf{Top:} the curves trace the top, middle, and bottom of the Static range, whose top flattens over the shaded recent-cutoff region. \textbf{Bottom:} both settings improve, Active about twice as fast, and the bands are bootstrap 95\% intervals. Slopes and the tests behind the quoted $p$ are in \S\ref{sec:scaling} and Appendix~\ref{app:stats}.}
\label{fig:convergence}
\end{figure}

Autonomous AI scientists are emerging as a new paradigm for scientific discovery, integrating literature review, hypothesis generation, experimentation, analysis, and manuscript preparation across the research lifecycle~\citep{lu2024aisci,gottweis2025coscientist,zheng2025automationautonomysurveylarge}.
Recent progress has been particularly pronounced in research execution, with agents increasingly capable of conducting long-horizon experimentation, optimizing algorithms, and generating scientific manuscripts with limited human intervention~\citep{jiang2025aideaidrivenexplorationspace,zhu2026ultralonghorizonagenticsciencecognitive,weng2025deepscientistadvancingfrontierpushingscientific}.
Scientific ideation remains a critical but underdeveloped foundation of this pipeline. It sets the direction and bounds what downstream execution can achieve, and the ideas current systems produce often fail to fulfill their apparent promise in practice~\citep{si2025ideationexecutiongapexecutionoutcomes}.
Evaluating and advancing autonomous AI scientists thus requires rigorous benchmarking of the ideation step itself.

A growing body of work already evaluates LLM ideation, and almost all of it shares a single elicitation protocol. The evaluation designer curates a fixed set of reference papers, the model reads them in one pass and proposes an idea, and a judge, usually another LLM, rates it on novelty, feasibility, and related axes \citep{si2024llmideas,guo2024ideabench,qiu2025aiideabench,wang2024scimon,li2024chainofideas,su2024virsci,hu2024nova,yang2025moosechem,zheng2023judge}. This static-observation paradigm has two limitations that motivate our benchmark. The first is a departure from how research actually works. A working scientist assembles their own evidence, querying the literature, deciding what to read next, following citation trails, and revising a forming idea against what turns up. The AI-scientist systems above operate the same way, interleaving retrieval with reasoning where a curated dossier would have to be read in one pass. Supplying the model with a reading list collapses that loop into a single synthesis step and measures how well the model recombines the references it was given. Whether it can explore the literature and gather its own evidence, the thing agentic deployment actually exercises, goes untested. The second limitation is that the paradigm loses resolution just where comparisons matter most. Reading a curated list and producing a fluent, plausible hypothesis has become a mature capability that recent models share, so scores saturate as models improve. Static quality keeps climbing with model recency while its top flattens across the recent frontier (Figure~\ref{fig:convergence}), and the strongest models pile against a ceiling that active exploration pushes past. In our evaluation the eight strongest models sit within 0.39 of a 1.53-point Static span, yet span 1.39 under active exploration (\S\ref{sec:landscape}).

\begin{figure*}[t]
\centering
\includegraphics[width=\textwidth]{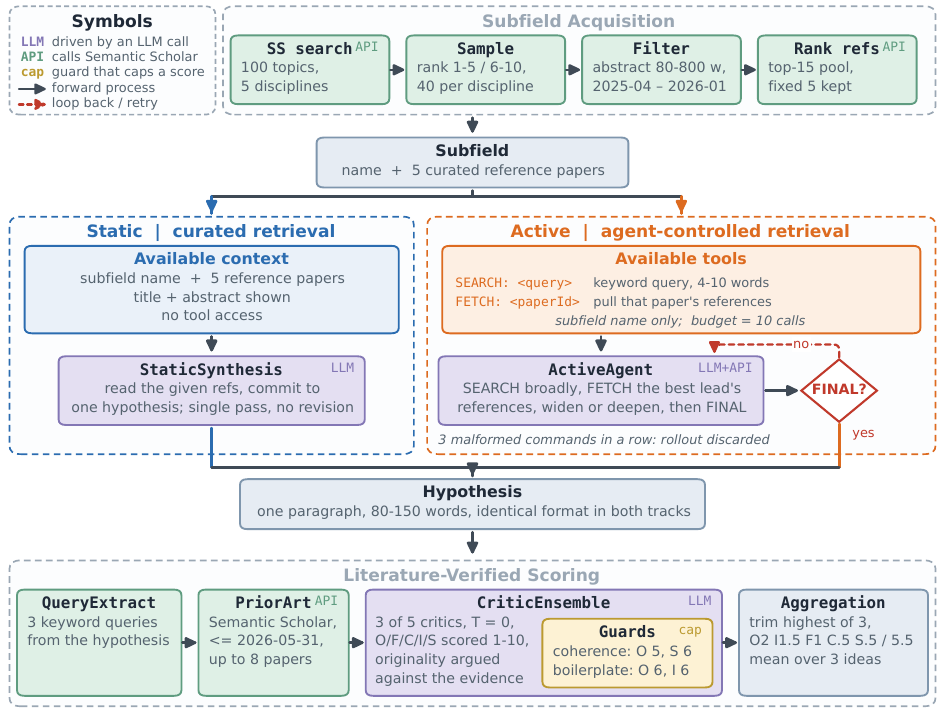}
\caption{\name{} in one view; schematic, no measured quantity appears. Subfields and their curated reference sets are built once (top), then one task set feeds two tracks matched on subfields, output format, length, and scoring. Their primary contrast is the literature-access protocol: Static is given its references and answers in a single pass, while Active is given only the subfield name plus \texttt{SEARCH}/\texttt{FETCH} under a $10$-call budget and loops until it emits \texttt{FINAL}, so it additionally entails multi-turn retrieval decisions and tool execution (\S\ref{sec:settings}). Both emit the same hypothesis into the same literature-verified critic, whose originality judgment is argued against prior art retrieved at judging time and capped by the coherence and boilerplate guards. Full protocol in \S\ref{sec:bench}.}
\label{fig:overview}
\end{figure*}

We introduce \name{}, a benchmark that makes the mode of literature access the experimental variable, contrasting \emph{static observation} of a curated literature pool with \emph{active exploration} in which the model gathers its own evidence. Every research subfield is posed under two matched settings that share prompts, output format, and scoring (Figure~\ref{fig:overview}). In the \textbf{Static} setting the model passively observes a curated set of reference papers for the subfield and synthesizes one 80--150-word hypothesis. In the \textbf{Active} setting the model receives only the subfield name together with a Semantic Scholar search tool under a fixed interaction budget, and must explore the literature itself before proposing in the identical format. An ensemble of three critics with literature verification scores each hypothesis on originality, feasibility, clarity, impact, and specificity under an originality-dominant weighting. The pipeline retrieves date-filtered prior art for the hypothesis and requires the critic to justify its originality score against that evidence rather than from its own memory. Using this setup we evaluate 35 models, 33 of them with matched Static--Active outputs. The primary roster holds 30 models, 28 of them matched and two Active-only, and is predominantly open-weight; Appendix~\ref{app:license} documents its single API-only Mistral entry. A further 5 closed-source models are held out for out-of-roster replication, and headline analyses use the 28 matched primary-roster models. Every model is run on 40 densely scored subfields drawn from 100 spanning computer science, physics, biology, chemistry, and medicine, with three hypotheses per model, subfield, and setting. The five disciplines serve as measurement coverage, and we advance no scientific claim within any of them.

Scoring the two settings side by side, we find that Active evaluation reveals capability headroom the Static protocol conceals (\S\ref{sec:results}). Along the knowledge-cutoff axis both settings improve, but Active improves about twice as fast, at $+1.16$ against $+0.54$ points per year, so the Active--Static gap widens as models become more capable. At the model level the gap is capability-gated. Static ability strongly predicts the Active gain at $r{=}{+}0.69$, the strongest models gain more than a full point while the weakest are hurt, and the gate replicates on five held-out closed-source models. A per-dimension decomposition locates the gain. Agent-controlled retrieval raises feasibility, clarity, and specificity, while measured originality stays flat at $-0.14$, which is not significant. Active exploration grounds hypotheses in evidence; under our critics it does not by itself make them more novel.

Scientists have long leaned on thought experiments to pressure-test an idea before committing to it. Motivated by this, we probe \textbf{Scientific World Modeling} (SWM), a generation-time refinement step. After the agent drafts a hypothesis, a structured thought experiment simulates the proposal, critiques its novelty and feasibility on separate channels while protecting a novel core, and feeds the outcome back for further retrieval and revision (\S\ref{sec:swm-main}). SWM helps the smaller backbones, with qwen-9b gaining $+0.61$ at $p{=}0.028$, and its benefit shrinks as the backbone strengthens until it disappears at the frontier, where the two deepseek-v4 backbones move $+0.14$ and $-0.06$ at $p{>}0.6$. In sampled reasoning traces the frontier backbones already run similar novelty and mechanism checks unprompted, an illustrative observation we do not treat as causal (Appendix~\ref{app:cot}). Explicit thought experiments would then help most where a model does not yet perform them on its own. We report SWM as an exploratory, capability-gated direction, since the pooled effect survives neither multiple-comparison correction nor a compute-matched baseline (\S\ref{sec:swm-main}).

These results recast scientific ideation as an agentic capability whose headroom static evaluation understates. Our contributions are
\textbf{(i)}~\name{}, a multidisciplinary benchmark that evaluates scientific ideation under matched Static and Active settings with literature-verified, originality-weighted critics, covering 33 matched models, five disciplines, and 40 densely scored subfields; \textbf{(ii)}~evidence that agentic evaluation produces top-end score separation where static scoring compresses, carrying $4.4\times$ the between-model score variance. The Active--Static gap is capability-gated at $r{=}{+}0.69$, replicates on a held-out closed-source family, and scales about twice as fast across model generations; \textbf{(iii)}~a per-dimension mechanism in which agent-controlled retrieval grounds feasibility, clarity, and specificity but leaves measured originality unchanged.

\begin{table*}[t]
\centering\small\setlength{\tabcolsep}{4.5pt}\renewcommand{\arraystretch}{1.0}
\begin{tabular}{l cccc rr}
\toprule
Benchmark / system & \makecell{Static\\Observation} & \makecell{Active\\Exploration} & \makecell{Literature-Grounded\\Judgement} & \makecell{Multi-\\Discipline} & \#\,Tasks & \makecell{\#\,Models\\tested} \\
\midrule
\multicolumn{7}{l}{\emph{Ideation benchmarks and human studies}} \\
IdeaBench \citep{guo2024ideabench}              & \cmark & \xmark & \xmark & \cmark & 2{,}374 & 7 \\
AI Idea Bench 2025 \citep{qiu2025aiideabench}   & \cmark & \xmark & \xmark & \xmark & 3{,}495 & 1 \\
LiveIdeaBench \citep{ruan2026liveideabench}     & \xmark & \xmark & \xmark & \cmark & 1{,}180 & $>$40 \\
MOOSE-Chem \citep{yang2025moosechem}            & \cmark & \xmark & \xmark & \xmark & 51      & 1 \\
HypoBench \citep{liu2025hypobench}              & \xmark & \xmark & \xmark & \pmark & 12      & 4 \\
\citet{si2024llmideas} (human study)            & \cmark & \pmark & \pmark & \xmark & 7       & 1 \\
\midrule
\multicolumn{7}{l}{\emph{Idea-generation systems}} \\
SciMON \citep{wang2024scimon}                   & \cmark & \pmark & \xmark & \pmark & ---     & 4 \\
Chain-of-Ideas \citep{li2024chainofideas}       & \xmark & \cmark & \xmark & \pmark & ---     & 1 \\
Nova \citep{hu2024nova}                         & \xmark & \cmark & \xmark & \xmark & 170     & --- \\
\midrule
\textbf{\name{} (ours)} & \cmark & \cmark & \cmark & \cmark & 100 & \textbf{35} \\
\bottomrule
\end{tabular}
\caption{\name{} versus prior ideation benchmarks, human studies, and idea-generation systems. \cmark~yes, \pmark~partial, \xmark~no. \textbf{Static Observation}: the model is given a reference set curated for the topic. \textbf{Active Exploration}: the model gathers its own evidence through tool calls it issues itself. \textbf{Literature-Grounded Judgement}: originality is scored against prior art retrieved at judging time instead of from the judge's memory. \textbf{Multi-Discipline}: the task set spans more than one scientific discipline. \textbf{\#\,Tasks} is the scale each paper reports, in that paper's own unit: idea instances for IdeaBench and AI Idea Bench, keywords for LiveIdeaBench, expert-annotated papers for MOOSE-Chem, tasks for HypoBench (over 194 datasets), research topics for \citet{si2024llmideas} (49 ideas per condition), and subfields for \name{}. \textbf{\#\,Models} counts distinct idea-generation backbones the paper evaluates, excluding judge and evaluator models, and \texttt{---} means the paper does not report the figure. \citet{si2024llmideas} judge by expert human review, which is a stronger standard than any automatic judge but is not literature-grounded in the sense of this column, hence \pmark. For \name{}, 33 of the 35 have matched Static--Active outputs. Discussion in \S\ref{sec:related}.}
\label{tab:comparison}
\end{table*}

\section{Related Work}\label{sec:related}

\paragraph{Autonomous AI scientists.} End-to-end ``AI scientist'' systems chain literature review, ideation, experimentation, and manuscript writing into a single research loop \citep{lu2024aisci,gottweis2025coscientist,weng2025deepscientistadvancingfrontierpushingscientific,jiang2025aideaidrivenexplorationspace}. Their advances concentrate on execution, where experiment design, coding, and writing can be scored against objective outcomes. The ideation that seeds the loop is still supplied largely by human insight, by pre-designed task specifications, or by a human in the loop who selects which hypotheses to pursue \citep{si2025ideationexecutiongapexecutionoutcomes,baek2024researchagent}. \name{} targets this under-measured step, asking whether a model can gather its own evidence and originate a hypothesis without that human scaffolding.

\paragraph{Ideation and hypothesis-generation benchmarks.} A parallel line evaluates ideation itself through human-preference studies \citep{si2024llmideas}, benchmarks that score ideas written from a curated reference set \citep{guo2024ideabench,qiu2025aiideabench,yang2025moosechem}, benchmarks that supply minimal context and score divergent thinking from a single keyword \citep{ruan2026liveideabench}, benchmarks that score hypotheses induced from data instead of literature \citep{liu2025hypobench}, and generation methods that structure retrieval or multi-role critique to raise novelty \citep{wang2024scimon,li2024chainofideas,hu2024nova,su2024virsci}. Two limitations recur across this line, and \name{} is built to remove them. At generation time the model either reads a reference list assembled for it or works from no literature at all, so it never decides what to read. At scoring time novelty is judged by an LLM with no access to the prior art it is judging against, so an idea reads as original whenever its precedents fall outside what the judge remembers, and LLM judges are known to carry systematic biases besides \citep{zheng2023judge,panickssery2024selfpref}. We contrast curated against agent-controlled retrieval under matched conditions (Table~\ref{tab:comparison}) and require every originality judgment to be argued against prior art retrieved at scoring time (\S\ref{sec:scoring}).

\section{\textsc{AgentIdeaBench}}\label{sec:bench}

\name{} turns scientific ideation into a matched-conditions comparison whose primary contrast is \emph{who controls the retrieval of supporting literature}. This section describes the tasks, the two tracks, and the literature-verified critic (Figure~\ref{fig:overview}). Table~\ref{tab:comparison} situates \name{} against prior ideation benchmarks and idea-generation systems. No other entry there combines all three of matched curated-against-agent-controlled retrieval, novelty scored against prior art retrieved at judging time, and a 35-model, five-discipline, cross-cutoff roster.

\subsection{Tasks and Subfields}\label{sec:tasks}
We curate 100 research subfields spanning computer science, physics, biology, chemistry, and medicine. They are chosen to cover both concept-driven fields, where a hypothesis is an argument, and experiment-heavy fields, where it is a protocol. Each subfield is a narrow, active research area such as a specific mechanism, material class, or model family, never a broad topic, so that a single focused hypothesis is a natural response and reference literature is well defined. For each subfield we assemble a reference set from Semantic Scholar, and for the scored subset a target ``ground-truth'' framing used only to build the Static context. The target literature is published after the knowledge cutoff of nearly all evaluated models, so by construction it is unseen at training time (Appendix~\ref{app:leakage}).

\subsection{Static and Active Settings}\label{sec:settings}
In \textbf{Static} mode, or \emph{curated retrieval}, the model receives the subfield name plus a fixed set of reference papers curated for it, each as a title and abstract, and must propose one novel, testable hypothesis as a single 80--150-word paragraph in first-person future tense, naming mechanisms, methods, and datasets. In \textbf{Active} mode, or \emph{agent-controlled retrieval}, the model receives \emph{only} the subfield name and a Semantic Scholar search/fetch tool under a fixed call budget, set past the point where added calls stop changing scores (Appendix~\ref{app:budget}). It must gather its own evidence, then propose in the identical format (Figure~\ref{fig:overview}, left). Both tracks share subfields, output format, budget on output length, and scoring, and both draw on external literature, so Static remains distinct from a closed-book recall condition. We grade a single free-text paragraph rather than a filled-in template, because structured multi-slot forms reward exhaustive box-filling and depress originality, whereas a paragraph forces the model to commit to one idea.

\paragraph{What the contrast isolates and what it bundles.} Active changes more than one thing relative to Static. Beyond gathering its own evidence, the agent decides what to read across turns and must emit well-formed tool calls within a budget. A track difference thus reflects agent-controlled retrieval as a bundle of retrieval control, multi-turn interaction, and tool-use competence, which is how we phrase every finding below. A recall-only track and a replay track bound this confound, and both are described with the rest of the setup in \S\ref{sec:controls}.

\subsection{Literature-Verified Scoring}\label{sec:scoring}
Each hypothesis is scored $1$--$10$ on \emph{originality, feasibility, clarity, impact, specificity} by an ensemble of open-weight critics, under an originality-dominant weighting. Originality carries the scientific core, while clarity and specificity are writing-quality axes that should not drive the ranking. The exact weights and aggregation are in \S\ref{sec:setup-eval}. Before scoring, the pipeline extracts queries from the hypothesis and retrieves date-filtered prior art from Semantic Scholar (Figure~\ref{fig:overview}, right). The critic receives this evidence block, must justify its originality score \emph{against it}, and is explicitly forbidden from overruling the retrieved evidence with its own memory. That closes the most damaging failure mode of LLM judges for ideation, in which an idea is called novel merely because its prior art lies outside the judge's training data. Coherence and boilerplate checks cap originality and specificity for keyword-stuffed or template-like text, so weak models cannot game the rubric with dense but vacuous proposals.

\subsection{Benchmark Scale and Validity}\label{sec:scale-validity}
Of the $100$ subfields, $40$ are densely scored with three hypotheses per model, subfield, and track cell to reduce sampling noise, amounting to roughly 21k critic calls. These are eight per discipline, chosen by a fixed deterministic rule given in Appendix~\ref{app:repro}. The full $100$ are generated in both tracks for coverage.

\paragraph{Human-landmark validation.} The score is an LLM judgment, so we anchor it against human ground truth. Landmark papers rewritten into the benchmark's format score $+1.00$ to $+2.24$ above the per-discipline all-model Active average, and a clean award-paper set, CORE-7, scores mean $7.71$, above every model's Active mean (Table~\ref{tab:leaderboard}). The scorer is the outcome of a seven-round calibration study (Appendix~\ref{app:critic}).

\section{Experimental Setup}\label{sec:setup}
We fix the model roster and generation settings (\S\ref{sec:setup-models}), the scoring protocol that turns critic judgments into a model score (\S\ref{sec:setup-eval}), and the controls and statistical tests that bound the findings (\S\ref{sec:controls}).

\subsection{Models and Generation}\label{sec:setup-models}
The primary roster is thirty models, $28$ of them with complete Static/Active pairs. The remaining two are reasoning-only models that reject the Static path and run Active-only. We additionally hold out five closed-source Gemini models for out-of-roster replication, giving $35$ evaluated models in all, $33$ of them with matched Static/Active outputs. Headline analyses use the $28$ matched primary-roster models, which are predominantly open-weight; Appendix~\ref{app:license} documents the single API-only Mistral entry. Models span knowledge cutoffs from $2023$ to $2026$, which lets us read ability against both model generation and knowledge recency. Generation uses temperature $0.7$ and seed $42$, with the Active agent given a $6000$-token limit and a $10$-call search/fetch budget. We fix that budget from a single-batch sweep over $\{1,2,5,10,15,20\}$ calls on five backbones, a Qwen3.5 size ladder and two other open-weight families. Paired tests find no backbone gaining between $5$ and $10$ calls (all $p\ge0.30$), while one still gains between $10$ and $20$: qwen3.5-9b by $+0.45$. A budget of $10$ is thus conservative for that one rather than flat for all five, and we keep it because doubling it doubles Active cost roster-wide for a gain confined to a single backbone. Realized utilization at budget $20$ spans $21$--$38\%$ and does not predict which backbones benefit. The sweep is reported in Appendix~\ref{app:budget}. The full roster with parameter counts, cutoff provenance, and exclusion rules is in Appendix~\ref{app:repro}, Table~\ref{tab:models}.

\subsection{Evaluation Protocol}\label{sec:setup-eval}
Three open-weight critics, \texttt{qwen3.6-plus}, \texttt{kimi-k2.6}, and \texttt{glm-5.1}, score each hypothesis on the five dimensions at temperature $0$. Per dimension we take a \emph{trimmed mean} that drops the single highest of the three critic scores and averages the rest, then combine dimensions under the originality-dominant weighting $O{:}2,\,I{:}1.5,\,F{:}1,\,C{:}0.5,\,S{:}0.5$, normalized by $5.5$. The three hypotheses in a cell are averaged, and a model's score is the mean over subfields. Before scoring, the pipeline retrieves date-filtered prior art for each hypothesis and requires the originality justification to cite it (\S\ref{sec:scoring}). Appendix~\ref{app:repro} gives the full aggregation, retrieval, and parsing-failure handling, and Appendix~\ref{app:judge} the judge-robustness checks over weighting, single judges, self-grading, and pairwise agreement.

\subsection{Controls and Statistical Tests}\label{sec:controls}
Two controls decompose the Active bundle. The \textbf{recall-only} track gives the model only the discipline and subfield name, with no curated references and no tools, so it must propose from parametric knowledge. Comparing Static, recall-only, and Active on their shared cells separates the effect of merely having references from the effect of agent-controlled retrieval (Appendix~\ref{app:recall}). The \textbf{replay} track refeeds each model, in the passive Static format, the exact references its own Active agent surfaced, reconstructed from stored Active telemetry and capped at the Static reference count. Holding delivery fixed at Static and varying only whether the references were designer-curated or agent-surfaced separates the content the agent gathers from the process of gathering it. The total gain Active$-$Static then splits additively into a content term Replay$-$Static and a process term Active$-$Replay (Appendix~\ref{app:replay}). Both controls' results are reported in \S\ref{sec:altrobust}.

\paragraph{Statistical tests.} Unless noted, effect sizes are paired differences over shared model-and-subfield cells with Wilcoxon signed-rank tests, and uncertainty is a bootstrap 95\% CI over $5000$ resamples with a fixed seed. The roster is family-heavy, since $9$ of $28$ models are qwen, so correlations are additionally cluster-bootstrapped over model families and checked leave-one-family-out. The track$\times$cutoff interaction is tested by within-model permutation. The capability gate is re-estimated on disjoint subfield halves, so that ability and gain carry independent scoring noise. Discrimination is measured by between-model variance ratios and a pooled-bootstrap-SE distinguishability criterion. Units, tests, effect sizes, and corrections for every headline claim are collected in Appendix~\ref{app:stats}.

\section{Results and Analysis}\label{sec:results}
Active evaluation reveals capability headroom that Static conceals, and that headroom is unevenly distributed across models. We first show that Active separates models Static compresses (\S\ref{sec:landscape}), then that the Active gain grows with model capability on two independent axes, static ability and knowledge-cutoff year (\S\ref{sec:gate}). A per-dimension decomposition places the gain in grounding rather than in measured originality (\S\ref{sec:mechanism}). We then separate reference presence from reference gathering, rule out remaining alternative explanations, and bound the findings' scope (\S\ref{sec:altrobust}). A separate section probes whether a generation-time procedure can add what retrieval does not supply (\S\ref{sec:swm-main}).\label{sec:analysis}

\subsection{Active Evaluation Improves Model Separation}\label{sec:landscape}
Table~\ref{tab:leaderboard} reports every paired model's Static and Active totals and their gap, the mean number of tool calls its Active rollouts issue, and its five Active per-dimension scores, ranked by Active total. Active quality spans $3.20$ to $6.33$ across three model generations, and it re-separates models that Static compresses near the top. The static protocol saturates at the frontier (Figure~\ref{fig:convergence}), while across the 28 primary-roster paired models the Active total carries $4.4\times$ the between-model variance of Static, and a per-model bootstrap-SE criterion distinguishes $62\%$ of top-half pairs against Static's $12\%$ (Appendix~\ref{app:stats}). Two patterns in the table set up the analysis that follows. The dimensions dissociate. Originality, impact, and the total climb toward the frontier, whereas feasibility is highest on weak models such as Llama-4 Maverick at $6.87$ and varies little with capability; we formalize that split in \S\ref{sec:mechanism}. The Active$-$Static gain $\Delta$ is also far from uniform. $19$ of 28 improve, the strongest by over a point, with GLM-5.1 at $+1.21$, while several weak models are hurt, with Gemma-2 27B at $-0.82$. We quantify this capability gate next. The five held-out Gemini models, marked $\dagger$, slot in among the strongest. A larger set of closed models scored outside this roster, and the capability gate recomputed on it, is reported in Appendix~\ref{app:fullroster}.

\begin{table*}[t]
\centering
\caption{\textbf{Model leaderboard.} Static (curated retrieval) and Active (agent-controlled retrieval) weighted totals for all 33 paired models, ranked by Active total; $\Delta{=}$Active$-$Static. \textbf{Turns} is the mean number of \texttt{SEARCH} and \texttt{FETCH} calls an Active rollout issues before its final synthesis, out of a budget of 10 (Appendix~\ref{app:budget}). It describes behavior rather than quality, so it is never bolded. The five Active per-dimension scores are each a 3-hypothesis $\times$ 40-subfield mean after dropping the most generous of three critics. Weighted total $O{:}2,I{:}1.5,F{:}1,C{:}0.5,S{:}0.5$ (normalized by $5.5$), under the \texttt{lit8d} critic (Appendix~\ref{app:critic}). Best open-weight value per column in \textbf{bold}. Held-out closed-source models are marked $\dagger$ and never bolded, and headline statistics use the 28 open-weight models. The dimensions do not move together. Originality and impact climb toward the frontier, whereas feasibility peaks on a weak model, Llama-4 Maverick at $6.87$, a dissociation formalized in \S\ref{sec:mechanism}.}
\label{tab:leaderboard}
\footnotesize\setlength{\tabcolsep}{3pt}\renewcommand{\arraystretch}{0.96}
\begin{tabular*}{\textwidth}{@{\extracolsep{\fill}} l rrr r rrrrr @{}}
\toprule
& \multicolumn{3}{c}{Weighted total} & & \multicolumn{5}{c}{Active per-dimension} \\
\cmidrule(lr){2-4} \cmidrule(lr){6-10}
Model & Static & Active & $\Delta$ & Turns & Orig. & Feas. & Clar. & Impact & Spec. \\
\midrule
GLM-5.1 & 5.13 & \textbf{6.33} & $\mathbf{+1.21}$ & 8.8 & \textbf{6.31} & 6.32 & \textbf{7.37} & \textbf{5.90} & \textbf{6.71} \\
Gemini-3 Flash$^\dagger$ & 5.59 & 6.29 & $+0.70$ & 8.5 & 6.33 & 6.16 & 7.12 & 5.92 & 6.60 \\
Gemini-3.5 Flash$^\dagger$ & 5.87 & 6.22 & $+0.35$ & 9.9 & 6.09 & 6.51 & 7.21 & 5.77 & 6.56 \\
Gemini-3.1 Pro$^\dagger$ & 5.63 & 6.20 & $+0.57$ & 7.5 & 6.20 & 6.14 & 7.14 & 5.83 & 6.47 \\
Kimi-K2.6 & 5.10 & 6.17 & $+1.08$ & 9.8 & 6.08 & 6.38 & 7.22 & 5.70 & 6.53 \\
Qwen3.5 397B & 5.06 & 5.95 & $+0.89$ & 8.9 & 5.79 & 6.28 & 6.81 & 5.57 & 6.19 \\
Kimi-K2.5 & 4.91 & 5.92 & $+1.02$ & 9.9 & 5.84 & 6.00 & 6.86 & 5.58 & 6.17 \\
DeepSeek-V4 Pro & \textbf{5.32} & 5.92 & $+0.60$ & 8.5 & 5.79 & 5.94 & 7.14 & 5.53 & 6.35 \\
MiMo-V2.5 Pro & 5.11 & 5.91 & $+0.81$ & 9.0 & 5.63 & 6.46 & 6.97 & 5.46 & 6.26 \\
GLM-4.6 & 4.83 & 5.78 & $+0.95$ & 6.9 & 5.64 & 6.14 & 6.84 & 5.31 & 5.99 \\
Qwen3.5 27B & 4.71 & 5.53 & $+0.82$ & 8.0 & 5.27 & 5.82 & 6.64 & 5.22 & 5.83 \\
DeepSeek-V4 Flash & 5.09 & 5.52 & $+0.42$ & 9.0 & 5.14 & 6.16 & 6.70 & 5.09 & 5.82 \\
Gemma-4 31B & 5.14 & 5.49 & $+0.35$ & 5.0 & 5.22 & 6.05 & 6.78 & 4.97 & 5.75 \\
MiMo-V2.5 & 4.90 & 5.46 & $+0.56$ & 7.8 & 5.14 & 6.05 & 6.35 & 5.17 & 5.55 \\
Mistral Medium 3.1 & 4.74 & 5.37 & $+0.62$ & 6.7 & 5.00 & 5.82 & 6.52 & 5.08 & 5.63 \\
MiniMax-M2.7 & 4.70 & 5.33 & $+0.62$ & 9.7 & 4.95 & 6.03 & 6.47 & 4.91 & 5.50 \\
DeepSeek-R1 & 4.86 & 5.30 & $+0.44$ & 6.0 & 5.08 & 5.50 & 6.49 & 4.95 & 5.68 \\
Mistral Small 2603 & 4.61 & 5.16 & $+0.55$ & 9.4 & 4.65 & 5.93 & 6.60 & 4.72 & 5.55 \\
Gemini-2.5 Flash$^\dagger$ & 5.10 & 5.05 & $-0.05$ & 7.7 & 4.53 & 6.07 & 6.01 & 4.78 & 4.92 \\
Qwen3 30B & 4.93 & 4.94 & $+0.01$ & 8.3 & 4.49 & 5.65 & 6.17 & 4.58 & 5.18 \\
Qwen3.5 9B & 4.23 & 4.87 & $+0.64$ & 6.9 & 4.75 & 4.90 & 6.03 & 4.53 & 5.13 \\
Gemma-3 27B & 4.68 & 4.82 & $+0.14$ & 6.2 & 4.16 & 6.34 & 6.09 & 4.21 & 5.02 \\
GLM-4.5 Air & 4.31 & 4.82 & $+0.51$ & 8.9 & 4.49 & 5.60 & 5.56 & 4.59 & 4.56 \\
Qwen3 Coder & 4.87 & 4.72 & $-0.15$ & 8.6 & 3.82 & 6.55 & 6.15 & 4.14 & 5.02 \\
Qwen3 32B & 4.53 & 4.44 & $-0.10$ & 3.1 & 4.13 & 5.04 & 5.31 & 4.15 & 4.46 \\
Gemini-2.5 Flash-Lite$^\dagger$ & 4.63 & 4.27 & $-0.37$ & 6.9 & 3.20 & 6.71 & 5.47 & 3.70 & 4.15 \\
Qwen3 8B & 4.24 & 4.21 & $-0.03$ & 4.3 & 3.84 & 5.10 & 4.95 & 3.90 & 4.13 \\
Qwen2.5 72B & 4.20 & 3.97 & $-0.23$ & 8.8 & 2.67 & 6.77 & 5.33 & 3.37 & 3.96 \\
Mistral Small 24B & 4.19 & 3.93 & $-0.26$ & 8.9 & 2.88 & 6.25 & 5.01 & 3.48 & 3.82 \\
Llama-4 Maverick & 4.34 & 3.84 & $-0.50$ & 4.5 & 2.63 & \textbf{6.87} & 4.77 & 3.23 & 3.47 \\
Qwen2.5 7B & 3.97 & 3.67 & $-0.30$ & 8.8 & 2.62 & 6.12 & 4.45 & 3.28 & 3.31 \\
Llama-3.1 8B & 3.80 & 3.54 & $-0.26$ & 6.7 & 2.65 & 5.66 & 4.11 & 3.29 & 3.06 \\
Gemma-2 27B & 4.02 & 3.20 & $-0.82$ & 4.1 & 2.26 & 5.60 & 3.90 & 2.75 & 2.75 \\
\bottomrule
\end{tabular*}
\end{table*}

\subsection{Active Gains Increase with Model Capability}\label{sec:gate}

\begin{figure}[t]
\centering
\includegraphics[width=\columnwidth]{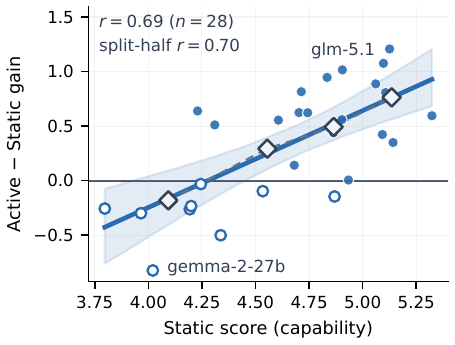}
\caption{The capability gate. Active$-$Static gain against static ability, one point per primary-roster paired model; filled markers are models Active helped, hollow markers those it hurt. White diamonds are the four static-ability quartile means, at $-0.18$, $+0.29$, $+0.49$, and $+0.76$. The band is a bootstrap 95\% interval for the fit. Correlations and robustness checks are in \S\ref{sec:gate} and Appendix~\ref{app:stats}.}
\label{fig:gate}
\end{figure}

Across the 28 matched primary-roster models the Active$-$Static gain is strongly correlated with static ability, at Pearson $r{=}{+}0.69$ and $p{<}10^{-4}$ (Figure~\ref{fig:gate}). The mean gain is a modest $+0.34$, with $19/28$ models improving at sign test $p{=}0.09$, so capability redistributes the gain more than it lifts every model. Grouped into static-ability quartiles, the weakest models lose $-0.18$ from being handed tools while the strongest gain $+0.76$. The correlation survives the family-heavy roster under a cluster bootstrap over nine families and leave-one-family-out, survives measuring ability and gain on disjoint subfield halves, and survives changes to the score weighting, the judge, and the removal of self-grading (Appendices~\ref{app:judge},~\ref{app:stats}). It replicates out-of-roster on the five held-out Gemini models at within-family $r{=}0.88$, and holds at $r{=}0.57$ when all $33$ are pooled.

\begin{figure}[t]
\centering
\includegraphics[width=\columnwidth]{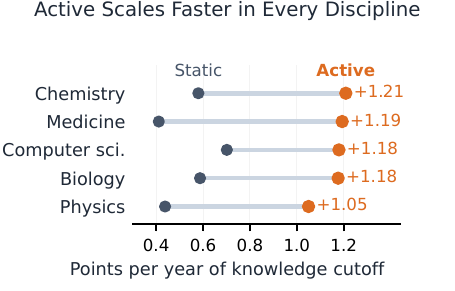}
\caption{Per-discipline scaling. Static and Active scores are regressed on knowledge-cutoff year separately within each discipline, giving one slope per track per discipline over the $28$ matched primary-roster models, of which $n{=}24$ carry a disclosed cutoff under Static and $n{=}26$ under Active. Active is steeper in all five. Static slopes span $+0.41$ to $+0.70$ and Active slopes $+1.05$ to $+1.21$, and their five-discipline means of $+0.54$ and $+1.16$ are the pooled slopes of \S\ref{sec:scaling}. Slopes are ordinary least squares, and per-discipline confidence intervals are not computed. Values are listed in Appendix~\ref{app:stats}.}
\label{fig:domainslopes}
\end{figure}

\label{sec:scaling}Model generation gives an independent axis, the knowledge-cutoff year, along which an advantage that compounds with capability should grow. Both settings improve, at different rates. Static quality rises $+0.54$/year, and its top compresses from the late-2024 cutoffs onward, where the eight strongest models span only $0.39$, so a static benchmark sees nothing past that plateau. Fitting the gap directly as a track$\times$cutoff interaction gives $+0.62$/year, 95\% CI $[0.41,0.94]$, permutation $p{<}10^{-3}$ (Appendix~\ref{app:stats}). Active is steeper in all five disciplines (Figure~\ref{fig:domainslopes}). The descriptive slopes are Static $+0.54$ against Active $+1.16$/year, a factor of ${\sim}2.1$ (Figure~\ref{fig:convergence}, bottom). General capability drives this rise, and memorization of the target papers does not (Appendix~\ref{app:leakage}). Ordered by rank or by cutoff, the Active advantage grows with capability.

\begin{figure}[t]
\centering
\includegraphics[width=\columnwidth]{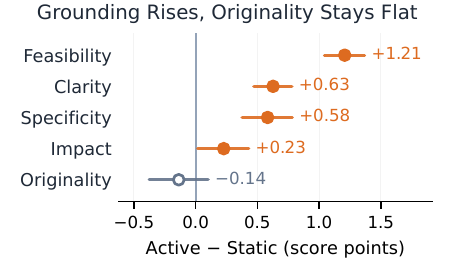}
\caption{Per-dimension Active$-$Static differences over the $28$ matched primary-roster models, with paired bootstrap 95\% CIs. Hollow markers are not significant under a Wilcoxon signed-rank test at $5\%$. See \S\ref{sec:mechanism}.}
\label{fig:perdim}
\end{figure}

\subsection{The Gain Lands in Grounding}\label{sec:mechanism}
The per-dimension Active$-$Static deltas locate the gain (Figure~\ref{fig:perdim}). Agent-controlled retrieval raises feasibility by $+1.21$, clarity by $+0.63$, specificity by $+0.58$, and impact by $+0.23$. The critic's originality score stays flat at $-0.14$, which is not significant, with $15/28$ models improving at sign test $p{=}0.85$. We read the pattern as grounding without added novelty. Retrieved evidence supplies concrete methods, datasets, and constraints that make a hypothesis more executable and precise, while under our evaluator it leaves the idea's measured originality where it was. The same reading accounts for the gate. A model that can already originate converts the added grounding into a higher weighted total, whereas a weaker model tends to convert the same grounding into competent restatement, which an originality-dominant weighting scores as a net loss. It also sets the target for \S\ref{sec:swm-main}, since a method that raises the total has to add the originality retrieval leaves untouched.

\subsection{Alternative Explanations and Robustness}\label{sec:altrobust}

\label{sec:recall}\textbf{Reference presence.} A recall-only control supplies the discipline and subfield name with neither curated references nor search tools, which separates having references from gathering them. Re-aggregated on the shared cells the ordering is Static${\approx}$Recall${<}$Active. Curated references do not significantly outscore closed-book recall, with Static$-$Recall at $-0.09$ and $p{=}0.11$, whereas agent-controlled retrieval does, with Active$-$Recall at $+0.23$ and $p{<}10^{-3}$ (Appendix~\ref{app:recall}). Reference presence alone adds little, and the gain requires the model to gather its own evidence.

\textbf{Content versus process.} A replay control refeeds each model in the passive Static format the very references its own Active agent surfaced. Passive replay preserves only a small and statistically insignificant part of the Active$-$Static gap, with Replay${-}$Static at $+0.08$ and $p{=}0.21$. Most of the difference stays with the interactive Active protocol, where Active${-}$Replay is $+0.26$ at $p{<}10^{-3}$ (Appendix~\ref{app:replay}). Handing a model its agent's own papers passively reproduces the Static score. Active bundles this process with multi-turn interaction and tool use (\S\ref{sec:settings}), which we do not further separate.

\label{sec:diversity}\label{sec:robust}\textbf{Further checks.} Additional analyses indicate that agent-controlled retrieval broadens exploration without that breadth explaining the score gain, that target-paper memorization does not explain the cutoff trend, that the capability gate is robust to critic and weighting choices, and that score discrimination is strongest in computer science and physics and weakest in dense-literature fields, a scope limit we carry into every cross-discipline claim. Full analyses are in Appendices~\ref{app:metric},~\ref{app:critic},~\ref{app:judge}, and~\ref{app:leakage}.


\section{Scientific World Modeling}\label{sec:swm-main}
Agent-controlled retrieval grounds a hypothesis without adding measured originality (\S\ref{sec:mechanism}), leaving open whether a generation-time procedure can supply what retrieval does not. Active inference gives that gap a vocabulary. An agent there selects actions that minimize expected free energy, which decomposes into a pragmatic term favoring preferred outcomes and an epistemic term favoring information gain \citep{friston2010fep,friston2015epistemic,friston2017processtheory}. Our Active track is epistemic action taken against the external literature, and \S\ref{sec:mechanism} finds its return landing on the pragmatic side of the score, in feasibility and precision. A generation-time loop targets the other term by querying the agent's own generative model in place of the literature, simulating what a proposal predicts before the agent commits to it, which is the role a world model plays in this account \citep{ha2018worldmodels,parr2022activeinference}. We report one attempt as a bounded, exploratory probe, and we take the framing as motivation rather than as a formal objective we optimize.

\paragraph{Method.} We extend the Active agent with a \texttt{SIMULATE} command that sends the current draft to a black-box \swm{}, which returns structured feedback carrying novelty and feasibility on separate channels, a protected ``novel core'' no edit may weaken, and a next action. At least one simulation is required before finalizing. We ablate five internal designs at a fixed interface. The strongest of them, \textbf{dynamic\,$+$\,hard}, convenes a dynamic expert panel per hypothesis and folds its reviews into a fixed novelty, feasibility, and referee assembly (Appendix~\ref{sec:improving}).

\begin{table}[t]
\centering\setlength{\tabcolsep}{6pt}\small\renewcommand{\arraystretch}{0.92}
\begin{tabular}{lrrrr}
\toprule
 & \multicolumn{2}{c}{qwen3.5} & \multicolumn{2}{c}{deepseek-v4} \\
\cmidrule(lr){2-3}\cmidrule(lr){4-5}
Design & 9b & 27b & flash & pro \\
\midrule
Baseline (active) & 4.82 & 5.46 & 5.25 & 5.61 \\
\quad SWM (multi-agent) & 5.19 & 5.72 & 5.24 & 5.63 \\
\quad SWM (dynamic-panel) & 5.03 & 5.74 & \textbf{5.53} & \textbf{5.67} \\
\quad$+$ constraint & \textbf{5.42} & \textbf{5.80} & 5.39 & 5.56 \\
\bottomrule
\end{tabular}
\caption{Closed-loop \swm{} against baseline, as absolute weighted scores by backbone. Appendices~\ref{sec:improving} and \ref{app:stats} give the further experiments and details.}
\label{tab:swm}
\end{table}

\paragraph{The gain depends on the backbone.} The dynamic\,$+$\,hard design has the largest pooled estimate at $+0.245$ (Table~\ref{tab:swm}), spread unevenly across backbones. It is largest on the smallest one, where qwen-9b gains $+0.61$ at $p{=}0.028$. It shrinks mid-range, where qwen-27b gains $+0.36$ at $p{=}0.071$. It is absent on the two frontier deepseek-v4 backbones, which move $+0.14$ and $-0.06$ at $p{>}0.6$. Where it helps, it lifts originality and feasibility together, by $+0.47$ and $+0.67$ on qwen. The value of an explicit thought experiment thus falls as the backbone strengthens, running opposite to the retrieval gate (\S\ref{sec:gate}). Four backbones make this a pattern rather than a tested law, and sampled chain-of-thought traces suggest why (Appendix~\ref{app:cot}).

\paragraph{Three controls bound the claim.} The pooled gain does not survive multiple-comparison correction at Holm $p{\geq}0.12$. It is also inseparable from an otherwise-identical variant that aggregates the same panel's reviews in free text, differing by $+0.03$ at $p{=}0.75$. A compute-matched best-of-3 baseline meets or exceeds it (Appendix~\ref{app:matched}), so even the nominal gain partly reflects extra inference-time compute. We report structured thought experiments as a capability-gated direction that remains unconfirmed.

\section{Conclusion}\label{sec:conclusion}

We introduced \name, which evaluates scientific ideation as an \emph{agentic} ability by separating the ideas a model forms from a handed-over reading list from those it gathers evidence for itself, both scored by a literature-verified critic. The value of agent-controlled retrieval is capability-gated. It grounds ideas by raising feasibility, clarity, and specificity while leaving their measured originality where it was, so it helps strong models and hurts weak ones. Active scaling continues past the local plateau where Static levels off. Retrieval equalizes idea diversity without equalizing quality, and a generation-time world model, offered as an exploratory extension, fails to beat a compute-matched baseline. We release the benchmark, the critic, and all scoring data.

\section*{Limitations}
\label{sec:limits}
\paragraph{The scorer is a model.} Every dimension score in this paper is an LLM judgment. We mitigate that with retrieved prior-art evidence, a critic ensemble, and human-landmark anchoring, and we report the validity checks in Appendices~\ref{app:critic} and~\ref{app:judge}, though mitigation falls short of validation. Originality is the weakest dimension in this respect. Our human-agreement checks support it least, and our headline claims turn on it, so every originality statement here should be read as a statement about measured originality under this critic. Discrimination also varies across the five disciplines, and where it is weakest we treat the rankings as soft. The validation we have not run, and the one we consider most important, is an expert study in which domain researchers rate idea novelty directly.

\paragraph{Active bundles more than retrieval control.} The Static/Active contrast is designed to vary who controls retrieval, but the Active condition also introduces multi-turn interaction and the competence to execute tools (\S\ref{sec:settings}). Our controls in \S\ref{sec:altrobust} and Appendices~\ref{app:recall} and~\ref{app:replay} narrow what the effect can be attributed to, without separating interaction from tool competence. Separating them would need a condition supplying agent-selected references without agent-driven turns, which we did not run. Results here should be read as effects of agent-controlled retrieval as a whole rather than of retrieval control in isolation.

\paragraph{The generation-time result is unconfirmed.} We report Scientific World Modeling as an exploratory probe, and the controls in \S\ref{sec:swm-main} and Appendix~\ref{sec:improving} do not establish it. Its apparent benefit is confined to part of a small set of backbones, and it clears neither the corrections nor the baselines we test it against. We also cannot rule out that a loop of this kind flatters the drafts it is given where it should stress-test them, and a placebo test with deliberately weak drafts is future work. Whether a differently designed loop would help stronger models remains open.

\section*{Ethics Statement}
\paragraph{Models, data, and artifacts.} All models we evaluate are publicly available and were accessed through a commercial API router under its terms of use. The literature side of the benchmark uses bibliographic metadata, titles, and abstracts returned by the Semantic Scholar API. We redistribute no full texts, and our released artifacts contain generated hypotheses, critic scores, and retrieval identifiers, never licensed publisher content.

\paragraph{Human participation.} The pairwise agreement study in Appendix~\ref{app:judge} was annotated by a small number of NLP researchers who took part voluntarily and unpaid, having been informed of the purpose of the task and of how their ratings would be used. Annotators saw only anonymized hypothesis text, with model identities and critic scores withheld, and we recorded nothing beyond their ratings, optional free-text comments, and timestamps. No demographic or otherwise personal data were collected, and no data concerning any third party were involved.

\paragraph{Risks and intended use.} The hypotheses in this benchmark are model output that no one has tested. They may be incorrect, unoriginal, or unsafe to act on, and this matters most in medicine, chemistry, and biology. Our scores are likewise LLM judgments, and they should not be used to gate funding, publication, or priority claims, nor to substitute for expert review. Both the models and the retrieval index over-represent English-language, well-indexed, highly cited work, so ideas and prior art outside that distribution may be systematically undervalued. We report where our critic discriminates weakest for that reason, in place of a uniform ranking. Automating ideation also makes it cheap to mass-produce plausible-sounding proposals, and we caution against deploying such systems in research or funding pipelines without expert oversight and fairness auditing.


\bibliography{custom}

\begin{thebibliography}{27}
\providecommand{\natexlab}[1]{#1}

\bibitem[{Baek et~al.(2024)Baek, Jauhar, Cucerzan, and
  Hwang}]{baek2024researchagent}
Jinheon Baek, Sujay~Kumar Jauhar, Silviu Cucerzan, and Sung~Ju Hwang. 2024.
\newblock {ResearchAgent}: Iterative research idea generation over scientific
  literature with large language models.
\newblock \emph{arXiv preprint arXiv:2404.07738}.

\bibitem[{Ethayarajh(2019)}]{ethayarajh2019anisotropy}
Kawin Ethayarajh. 2019.
\newblock How contextual are contextualized word representations? comparing the
  geometry of {BERT}, {ELMo}, and {GPT}-2 embeddings.
\newblock In \emph{Proceedings of EMNLP-IJCNLP}.

\bibitem[{Friedman and Dieng(2023)}]{friedman2023vendi}
Dan Friedman and Adji~Bousso Dieng. 2023.
\newblock The {Vendi} score: A diversity evaluation metric for machine
  learning.
\newblock \emph{Transactions on Machine Learning Research (TMLR)}.

\bibitem[{Friston(2010)}]{friston2010fep}
Karl Friston. 2010.
\newblock The free-energy principle: a unified brain theory?
\newblock \emph{Nature Reviews Neuroscience}, 11(2):127--138.

\bibitem[{Friston et~al.(2017)Friston, FitzGerald, Rigoli, Schwartenbeck, and
  Pezzulo}]{friston2017processtheory}
Karl Friston, Thomas FitzGerald, Francesco Rigoli, Philipp Schwartenbeck, and
  Giovanni Pezzulo. 2017.
\newblock Active inference: A process theory.
\newblock \emph{Neural Computation}, 29(1):1--49.

\bibitem[{Friston et~al.(2015)Friston, Rigoli, Ognibene, Mathys, FitzGerald,
  and Pezzulo}]{friston2015epistemic}
Karl Friston, Francesco Rigoli, Dimitri Ognibene, Christoph Mathys, Thomas
  FitzGerald, and Giovanni Pezzulo. 2015.
\newblock Active inference and epistemic value.
\newblock \emph{Cognitive Neuroscience}, 6(4):187--214.

\bibitem[{Gottweis et~al.(2025)Gottweis, Weng, Daryin, Tu, Sirkovic,
  Myaskovsky, Glowaty, Weissenberger, Orlandi, Popovici, Palepu, Rong, Tanno,
  Saab, Zhang, Blum, Carroll, Kulkarni, Tomasev, Zverinski, Rendulic, Vedadi,
  Hasler, Rimanic, Boia, Budiselic, Feinstein, Bellaiche, Sheffer, Freyberg,
  Ratcliff, Bertolli, Chou, Hassidim, Gokturk, Vahdat, Guan, Dhillon, Vaishnav,
  Lee, Costa, Pened{\'e}s, Peltz, Matias, Manyika, Hassabis, Xu, Kohli,
  Pawlosky, Karthikesalingam, and Natarajan}]{gottweis2025coscientist}
Juraj Gottweis, Wei-Hung Weng, Alexander Daryin, Tao Tu, Petar Sirkovic, Artiom
  Myaskovsky, Grzegorz Glowaty, Felix Weissenberger, Alessio Orlandi, Dan
  Popovici, Anil Palepu, Keran Rong, Ryutaro Tanno, Khaled Saab, Fan Zhang,
  Jacob Blum, Andrew Carroll, Kavita Kulkarni, Nenad Tomasev, and 32 others.
  2025.
\newblock Accelerating scientific discovery with {Co-Scientist}.
\newblock \emph{arXiv preprint arXiv:2502.18864}.

\bibitem[{Guo et~al.(2024)Guo, Shariatmadari, Xiong, Huang, Xie, Bekiranov, and
  Zhang}]{guo2024ideabench}
Sikun Guo, Amir~Hassan Shariatmadari, Guangzhi Xiong, Albert Huang, Eric Xie,
  Stefan Bekiranov, and Aidong Zhang. 2024.
\newblock {IdeaBench}: Benchmarking large language models for research idea
  generation.
\newblock \emph{arXiv preprint arXiv:2411.02429}.

\bibitem[{Ha and Schmidhuber(2018)}]{ha2018worldmodels}
David Ha and J{\"u}rgen Schmidhuber. 2018.
\newblock Recurrent world models facilitate policy evolution.
\newblock In \emph{Advances in Neural Information Processing Systems}.

\bibitem[{Hu et~al.(2024)Hu, Fu, Wang, Wang, Li, Xu, Lu, Jin, Pan, and
  Lan}]{hu2024nova}
Xiang Hu, Hongyu Fu, Jinge Wang, Yifeng Wang, Zhikun Li, Renjun Xu, Yu~Lu,
  Yaochu Jin, Lili Pan, and Zhenzhong Lan. 2024.
\newblock Nova: An iterative planning and search approach to enhance novelty
  and diversity of {LLM} generated ideas.
\newblock \emph{arXiv preprint arXiv:2410.14255}.

\bibitem[{Jiang et~al.(2025)Jiang, Schmidt, Srikanth, Xu, Kaplan, Jacenko, and
  Wu}]{jiang2025aideaidrivenexplorationspace}
Zhengyao Jiang, Dominik Schmidt, Dhruv Srikanth, Dixing Xu, Ian Kaplan, Deniss
  Jacenko, and Yuxiang Wu. 2025.
\newblock \href {https://arxiv.org/abs/2502.13138} {Aide: Ai-driven exploration
  in the space of code}.
\newblock \emph{Preprint}, arXiv:2502.13138.

\bibitem[{Li et~al.(2024)Li, Xu, Guo, Zhao, Li, Yuan, Zhang, Jiang, Xin, Dang,
  Zhao, Rong, Feng, and Bing}]{li2024chainofideas}
Long Li, Weiwen Xu, Jiayan Guo, Ruochen Zhao, Xingxuan Li, Yuqian Yuan, Boqiang
  Zhang, Yuming Jiang, Yifei Xin, Ronghao Dang, Deli Zhao, Yu~Rong, Tian Feng,
  and Lidong Bing. 2024.
\newblock Chain of ideas: Revolutionizing research via novel idea development
  with {LLM} agents.
\newblock \emph{arXiv preprint arXiv:2410.13185}.

\bibitem[{Liu et~al.(2025)Liu, Huang, Hu, Zhou, and Tan}]{liu2025hypobench}
Haokun Liu, Sicong Huang, Jingyu Hu, Yangqiaoyu Zhou, and Chenhao Tan. 2025.
\newblock {HypoBench}: Towards systematic and principled benchmarking for
  hypothesis generation.
\newblock \emph{arXiv preprint arXiv:2504.11524}.

\bibitem[{Lu et~al.(2024)Lu, Lu, Lange, Foerster, Clune, and Ha}]{lu2024aisci}
Chris Lu, Cong Lu, Robert~Tjarko Lange, Jakob Foerster, Jeff Clune, and David
  Ha. 2024.
\newblock The {AI} {Scientist}: Towards fully automated open-ended scientific
  discovery.
\newblock \emph{arXiv preprint arXiv:2408.06292}.

\bibitem[{Panickssery et~al.(2024)Panickssery, Bowman, and
  Feng}]{panickssery2024selfpref}
Arjun Panickssery, Samuel~R. Bowman, and Shi Feng. 2024.
\newblock {LLM} evaluators recognize and favor their own generations.
\newblock In \emph{Advances in Neural Information Processing Systems
  (NeurIPS)}.

\bibitem[{Parr et~al.(2022)Parr, Pezzulo, and
  Friston}]{parr2022activeinference}
Thomas Parr, Giovanni Pezzulo, and Karl~J. Friston. 2022.
\newblock \emph{Active Inference: The Free Energy Principle in Mind, Brain, and
  Behavior}.
\newblock MIT Press.

\bibitem[{Qiu et~al.(2025)Qiu, Zhang, Xu, Li, Song, Wang, and
  Zhang}]{qiu2025aiideabench}
Yansheng Qiu, Haoquan Zhang, Zhaopan Xu, Ming Li, Diping Song, Zheng Wang, and
  Kaipeng Zhang. 2025.
\newblock {AI Idea Bench 2025}: {AI} research idea generation benchmark.
\newblock \emph{arXiv preprint arXiv:2504.14191}.

\bibitem[{Ruan et~al.(2026)Ruan, Wang, Hong, Wang, Liu, and
  Sun}]{ruan2026liveideabench}
Kai Ruan, Xuan Wang, Jixiang Hong, Peng Wang, Yang Liu, and Hao Sun. 2026.
\newblock Evaluating {LLMs}' divergent thinking capabilities for scientific
  idea generation with minimal context.
\newblock \emph{Nature Communications}.
\newblock ArXiv:2412.17596.

\bibitem[{Si et~al.(2025)Si, Hashimoto, and
  Yang}]{si2025ideationexecutiongapexecutionoutcomes}
Chenglei Si, Tatsunori Hashimoto, and Diyi Yang. 2025.
\newblock \href {https://arxiv.org/abs/2506.20803} {The ideation-execution gap:
  Execution outcomes of llm-generated versus human research ideas}.
\newblock \emph{Preprint}, arXiv:2506.20803.

\bibitem[{Si et~al.(2024)Si, Yang, and Hashimoto}]{si2024llmideas}
Chenglei Si, Diyi Yang, and Tatsunori Hashimoto. 2024.
\newblock Can {LLMs} generate novel research ideas? a large-scale human study
  with 100+ {NLP} researchers.
\newblock \emph{arXiv preprint arXiv:2409.04109}.

\bibitem[{Su et~al.(2024)Su, Chen, Tang, Yin, Zheng, Li, Qi, Wu, Li, Ouyang,
  Torr, Zhou, and Dong}]{su2024virsci}
Haoyang Su, Renqi Chen, Shixiang Tang, Zhenfei Yin, Xinzhe Zheng, Jinzhe Li,
  Biqing Qi, Qi~Wu, Hui Li, Wanli Ouyang, Philip Torr, Bowen Zhou, and Nanqing
  Dong. 2024.
\newblock Many heads are better than one: Improved scientific idea generation
  by a {LLM}-based multi-agent system.
\newblock \emph{arXiv preprint arXiv:2410.09403}.

\bibitem[{Wang et~al.(2024)Wang, Downey, Ji, and Hope}]{wang2024scimon}
Qingyun Wang, Doug Downey, Heng Ji, and Tom Hope. 2024.
\newblock {SciMON}: Scientific inspiration machines optimized for novelty.
\newblock In \emph{Proceedings of the 62nd Annual Meeting of the Association
  for Computational Linguistics (ACL)}.

\bibitem[{Weng et~al.(2025)Weng, Zhu, Xie, Sun, Lin, Liu, and
  Zhang}]{weng2025deepscientistadvancingfrontierpushingscientific}
Yixuan Weng, Minjun Zhu, Qiujie Xie, Qiyao Sun, Zhen Lin, Sifan Liu, and Yue
  Zhang. 2025.
\newblock \href {https://arxiv.org/abs/2509.26603} {Deepscientist: Advancing
  frontier-pushing scientific findings progressively}.
\newblock \emph{Preprint}, arXiv:2509.26603.

\bibitem[{Yang et~al.(2025)Yang, Liu, Gao, Xie, Li, Ouyang, Poria, Cambria, and
  Zhou}]{yang2025moosechem}
Zonglin Yang, Wanhao Liu, Ben Gao, Tong Xie, Yuqiang Li, Wanli Ouyang, Soujanya
  Poria, Erik Cambria, and Dongzhan Zhou. 2025.
\newblock {MOOSE-Chem}: Large language models for rediscovering unseen
  chemistry scientific hypotheses.
\newblock In \emph{International Conference on Learning Representations
  (ICLR)}.

\bibitem[{Zheng et~al.(2023)Zheng, Chiang, Sheng, Zhuang, Wu, Zhuang, Lin, Li,
  Li, Xing, Zhang, Gonzalez, and Stoica}]{zheng2023judge}
Lianmin Zheng, Wei-Lin Chiang, Ying Sheng, Siyuan Zhuang, Zhanghao Wu, Yonghao
  Zhuang, Zi~Lin, Zhuohan Li, Dacheng Li, Eric~P Xing, Hao Zhang, Joseph~E
  Gonzalez, and Ion Stoica. 2023.
\newblock Judging {LLM}-as-a-judge with {MT}-bench and chatbot arena.
\newblock In \emph{Advances in Neural Information Processing Systems
  (NeurIPS)}.

\bibitem[{Zheng et~al.(2025)Zheng, Deng, Tsang, Wang, Bai, Wang, and
  Song}]{zheng2025automationautonomysurveylarge}
Tianshi Zheng, Zheye Deng, Hong~Ting Tsang, Weiqi Wang, Jiaxin Bai, Zihao Wang,
  and Yangqiu Song. 2025.
\newblock \href {https://arxiv.org/abs/2505.13259} {From automation to
  autonomy: A survey on large language models in scientific discovery}.
\newblock \emph{Preprint}, arXiv:2505.13259.

\bibitem[{Zhu et~al.(2026)Zhu, Cai, Liu, Zheng, Wang, Ye, Zhang, Zhang, E,
  Chen, and Wang}]{zhu2026ultralonghorizonagenticsciencecognitive}
Xinyu Zhu, Yuzhu Cai, Zexi Liu, Bingyang Zheng, Cheng Wang, Rui Ye, Yuzhi
  Zhang, Linfeng Zhang, Weinan E, Siheng Chen, and Yanfeng Wang. 2026.
\newblock \href {https://arxiv.org/abs/2601.10402} {Toward ultra-long-horizon
  agentic science: Cognitive accumulation for machine learning engineering}.
\newblock \emph{Preprint}, arXiv:2601.10402.

\end{thebibliography}

\appendix

\section{Metric critique: centroid cosine cannot measure exploration}\label{app:metric}
The Static and Active reference sets for the same subfield have mean Jaccard overlap $0.032$, so the papers are nearly disjoint, yet their mean embedding-centroid cosine is $0.868$ over $n{=}1471$ pairs. That apparent similarity is a floor artifact.
(i)~Randomly halving a single topic's papers into two sets already yields cosine $0.884{\pm}0.041$, which is the same-topic floor.
(ii)~Cross-topic pairs score $0.206$.
(iii)~Replacing a growing fraction of a set's papers barely moves cosine while Jaccard collapses (Table~\ref{tab:sens}).
Centroid cosine within a subfield cannot support claims about retrieval similarity or diversity. Its narrow range stems from embedding anisotropy \citep{ethayarajh2019anisotropy} and from mean-pooling acting as a low-pass filter.

\begin{table}[h]
\centering\small
\begin{tabular}{lcccccc}
\toprule
frac.\ replaced & 0 & .2 & .4 & .6 & .8 & 1.0 \\
\midrule
centroid cos & 1.00 & .977 & .953 & .933 & .908 & .884 \\
Jaccard      & 1.00 & .667 & .429 & .250 & .111 & .000 \\
\bottomrule
\end{tabular}
\caption{Sensitivity analysis replacing a fraction of one reference set's papers with other same-topic papers ($n{=}100$ topics). Cosine spans $1.0{\to}0.88$ while true overlap spans $1.0{\to}0.0$.}
\label{tab:sens}
\end{table}

Set-level metrics separate the regimes cleanly. The mutual-$k$NN cross-set fraction is $0.331$ observed against $0.519$ for a random same-topic split, where 0.5 means fully interleaved. Energy distance is $0.143$ against $0.087$. The Vendi Score \citep{friedman2023vendi} effective-distinct-count is Static $4.08$ and Active $8.31$, roughly ${\sim}2\times$ more diverse, with the union at $9.45$, above either alone. The two searches contribute genuinely different content despite the $0.87$ cosine, and they differ in kind as well as in identity. The Active agent's self-retrieved papers are markedly newer and less canonical than the curated Static bibliography, with median publication year $2023$ against $2020$ and median citation count $16$ against $2{,}207$. The agent reaches current, long-tail work where the curated set holds the field's most-cited anchors. At the idea level each hypothesis stays about equally close to the literature regardless of who retrieved it. The idea--reference centroid cosine is $0.504$ for a Static idea against Static refs, $0.489$ for an Active idea against Static refs, and $0.487$ for an Active idea against Active refs. Those three pairings are nearly identical, which places the track difference in what is retrieved rather than in how tightly the idea tracks its references.

\section{Judge validity and robustness}\label{app:judge}
\begin{figure}[t]
\centering
\includegraphics[width=\columnwidth]{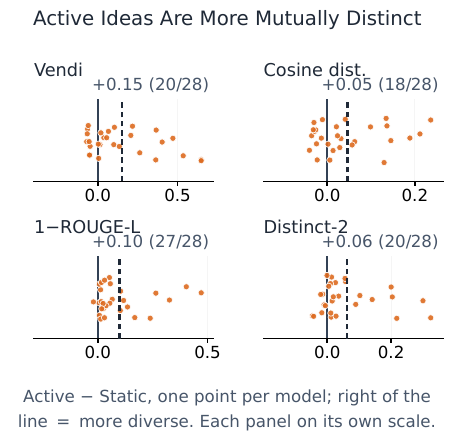}
\caption{Within-cell idea diversity, shown as the Active$-$Static difference on four metrics, one point per paired model at $n{=}28$, with the mean marked. Points right of zero are more diverse. Self-ROUGE-L is flipped to distinctness, computed as $1-$overlap, and each panel carries its own scale. Pooled cell-level statistics are in the paragraph above.}
\label{fig:diversity}
\end{figure}

\paragraph{Weight sensitivity.} The capability-gate correlation holds under every weighting we tried, namely equal weights, originality-only, and dropping clarity and specificity. Across these, $r$ ranges $0.65$--$0.79$, all at $p{<}10^{-3}$, and model rankings correlate at Spearman $\geq 0.988$ with the default weighting.
\paragraph{Judge ensemble.} The gate holds under each single judge, and inter-judge ranking agreement is $0.95$--$0.99$. Removing all cases where a judge grades its own model family leaves results and rankings essentially unchanged.
\paragraph{Pairwise judge agreement.} As an independent check beyond absolute scores, we ran a pairwise protocol. For $100$ hypothesis pairs, three critics each choose the better hypothesis in both presentation orders. Binary A-vs-B inter-judge agreement is Fleiss $\kappa{=}0.97$, or $\kappa{=}0.81$ ternary once ties are allowed. Position bias, measured as the fraction of order-flips that change a verdict, is $5.7\%$, and the LLM majority agrees with any single critic $93\%$ of the time at Cohen $\kappa{=}0.86$. This pairwise agreement complements the $0.95$--$0.99$ rank agreement of the absolute scorer. A preliminary human study over $100$ computer-science pairs used multiple annotators and produced human$-$human Fleiss $\kappa{=}0.44$ binary, reflecting genuine disagreement on close pairs. The critic agrees with the human majority on $77\%$ of pairs at Cohen $\kappa{=}0.55$, rising to $96\%$ and $\kappa{=}0.93$ on the $27$ large-margin pairs. The scorer is thus corroborated on clear cases, and close-call originality is where human and model judgment diverge most.
\paragraph{Human-landmark anchors.} Landmark papers rewritten into the benchmark's hypothesis format outscore the per-discipline all-model Active average by $+1.00$ to $+2.24$ in all five disciplines. Against the single strongest model the gap narrows sharply in the experiment-heavy, dense-literature fields of biology, chemistry, and medicine. We treat those disciplines' scores as soft rankings and consider computer science and physics most reliable.
\paragraph{Within-cell idea diversity.} Full within-cell diversity statistics over $n{=}2793$ paired cells are Vendi $1.943{\to}2.092$ with CI $[+0.134,+0.165]$, pairwise cosine distance $0.344{\to}0.390$, self-ROUGE-L $0.298{\to}0.199$, and distinct-2 $0.820{\to}0.880$, all at Wilcoxon $p{<}10^{-6}$. Static Vendi correlates with capability at $r{=}{+}0.442$ and $p{=}0.018$, while the Active$-$Static Vendi gain anti-correlates at $r{=}{-}0.472$ and $p{=}0.011$. Figure~\ref{fig:diversity} plots the four within-cell metrics.

\section{Prompts and protocol}\label{app:prompts}

\paragraph{Active protocol.} The agent emits exactly one plain-text command per turn: \texttt{SEARCH:\,<query>} (returns up to 10 papers with title$+$abstract$+$id), \texttt{FETCH:\,<paperId>} (returns the paper's top-15 references), and \texttt{FINAL:\,<hypothesis>} (80--150 words, first-person future tense, naming mechanisms/methods/datasets), under a fixed tool-call budget. The \swm{} variant adds \texttt{SIMULATE:\,<draft>} and requires at least one simulation before \texttt{FINAL}.

\paragraph{\swm{} feedback schema.} The schema is a single JSON object holding \texttt{mechanism\_consistency\{verdict, explanation, grounded\}}; \texttt{thought\_experiment\{setup, predicted\_outcome, confidence, calibration\_note\}}, whose calibration note marks the prediction as a plausibility signal and never as ground truth, with confidence never gating acceptance; \texttt{novel\_core}, one sentence that must survive any edit; \texttt{dimension\_weakness}, covering five dimensions at severity 0--3 with novelty and feasibility flagged as separate channels; \texttt{feasibility\_repair}, a concrete edit that must not weaken \texttt{novel\_core}; \texttt{knowledge\_gap}; and \texttt{next\_probe\{action $\in$ SEARCH$|$REFINE$|$ACCEPT, query, rationale\}}. A deterministic referee picks \texttt{next\_probe}. It searches if the mechanism is inconsistent or a fact is missing, refines if any severity is $\geq 2$, and accepts otherwise.

\paragraph{S4b hard-constrained assembly.} In step one a LEAD planner convenes 2--4 bespoke experts for the hypothesis at hand, specifying role, focus, and an optional retrieval query, with at least one novelty-focused and one feasibility-focused. In step two each expert reviews independently, returning a verdict, key point, per-dimension concerns, and one suggestion, with retrieved prior art if requested. In step three the panel's reviews become evidence for a fixed chain. A Novelty-scout names \texttt{novel\_core} and rates originality against the literature. A Feasibility-critic is then given \texttt{novel\_core} with the instruction ``DO NOT WEAKEN'' and rates feasibility, clarity, impact, and specificity before writing the repair. The deterministic referee assembles the schema and picks the next action. S4 is identical through step two and replaces step three with a single free-form LEAD synthesis prompt, the only difference between the two.

\paragraph{Critic.} Each hypothesis is scored by three open-weight critics on the five dimensions with anchored 1--10 rubrics. The prompt includes a retrieved prior-art evidence block and requires the originality justification to cite it, and coherence and boilerplate checks cap originality and specificity for keyword-stuffed or template-like text. Full verbatim prompts ship with the code release.

\paragraph{S5 gated variant.} S5 is S4b with one change. The Feasibility-critic's repair is suppressed when feasibility is already strong at severity $<2$, so an already-feasible draft passes through unedited where it would otherwise be forced through a lateral revision. The per-dimension diagnosis motivates the change, since S4b's always-on repair costs originality on saturated backbones, with deepseek originality at $-0.22$. A small pilot at $n{=}31$/$17$ suggested S5 beat S4b, and at the full sample of $n{=}45$ per deepseek backbone the effect reversed. S5$-$S4b is $-0.04$ on v4-flash and $-0.16$ on v4-pro, with originality still negative at $-0.42$ on v4-pro. Gating does not recover the saturated backbones, and the pilot's positive sign was a small-sample artifact. We keep this as a documented negative result.

\section{Critic design and calibration}\label{app:critic}
The literature-verified critic is the result of a seven-round calibration study, which we summarize here because the benchmark's validity rests on it.

\paragraph{The ceiling is partly the scorer's.} A naive originality-weighted critic scores real, highly-cited landmark hypotheses, rewritten to the benchmark format, at mean $4.78$. That is below the all-model average of $5.20$, with $0\%$ exceeding $6.5$. Fifteen ICLR-2026 Oral ideas used as anchors scored $5.54$, only $+0.34$ over models, with $0\%$ above $6.5$. The apparent ``$6.x$ wall'' was thus substantially a scoring artifact rather than a model ceiling.

\paragraph{Two failure modes.} The first is hindsight and knowledge asymmetry. A landmark's idea is in the critic's training data, so it is recognized as ``known'' and its originality suppressed to $\sim$$5$, while a model's fluent recombination has no findable prior art and is over-credited at $\sim$$8$. The second is detail-density gaming, where clarity and specificity reward named entities and precise restatements of already-known mechanisms, a bias strong models exploit.

\paragraph{Fixes in rubric \texttt{lit8d}.} We (a)~retrieve date-filtered prior art per hypothesis and require originality to be argued against that evidence, so an idea whose prior art is retrieved cannot be called novel; (b)~redefine specificity and clarity as precision of the new contribution rather than of restated background; (c)~calibrate feasibility for theory and algorithm ideas, where a decisive test suffices and lacking wet-lab work carries no penalty; and (d)~tie impact to evidence tiers, so that restating a known mechanism caps impact. We also recruited a clean anchor set, \textbf{CORE-7}, built from NeurIPS-2025 and ICLR-2026 award papers rewritten as explicit-increment claims. Under \texttt{lit8d}, CORE-7 scores mean $7.71$ with a peak of $8.5$ and $6/7\geq7.5$. That is $+1.38$ above the best leaderboard Active mean of $6.33$ from glm-5.1, and $+1.92$ above deepseek-v4-pro, the top model by Static score, on the shared anchor subset at $5.79$, or $5.92$ over the full 40. Three weak-control generators meanwhile stay at $1.80$, so the rubric discriminates where a uniformly inflating one would not. Robustness to held-out judges, at Spearman $0.805$ between selection and held-out critics, rules out same-family self-grading.

\paragraph{Coherence/boilerplate caps.} A final rubric guard caps originality and specificity for incoherent, keyword-stuffed, or template-like proposals, so weak models cannot inflate scores with dense but vacuous text. Ablating these caps and rescoring four backbones changes mean Originality by under $0.05$ on three of them, namely gemma-4-31b at $-0.02$, llama-3.1-8b at $-0.04$, and mistral-7b-v0.1 at $-0.03$, and by $-0.39$ on qwen3.5-9b. The caps rarely bind on coherent proposals, so they guard against gaming without materially distorting normal scores.

\begin{figure}[t]
\centering
\includegraphics[width=\columnwidth]{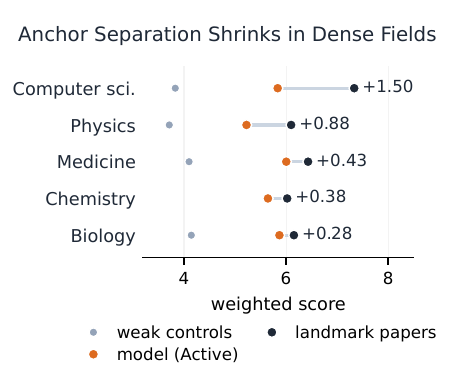}
\caption{Per-discipline critic validity, covering weak-probe control generators, the strongest model's Active ideas, and rewritten landmark papers, with the anchor$-$model gap. Chemistry has no weak-control items, so that marker is absent. Discussion appears under Domain scope in \S\ref{sec:robust} and below.}
\label{fig:domvalid}
\end{figure}

\paragraph{Domain validity.} The rubric separates award papers from the strongest model cleanly in concept-driven fields but not in experiment-heavy ones. Top-10 anchor$-$model gaps are computer science $+1.50$, physics $+0.88$, medicine $+0.43$, chemistry $+0.38$, and biology $+0.28$ (Figure~\ref{fig:domvalid}). Restricted to originality they are CS $+2.14$, physics $+1.01$, chemistry $+0.74$, medicine $+0.60$, and biology $+0.28$. In dense-literature fields a real mechanistic advance is retrieved as ``same line'' and a model's plausibly-worded mechanism also lands mid-scale, so scores converge. We treat CS and physics as the reliable regimes and biology, chemistry, and medicine as soft rankings. The limit is one of scope. Weak-model discrimination holds in all five disciplines, with controls at $1.80<$ weak $3.6$--$4.5<$ strong $5.4$--$6.0<$ CS anchors $7.7$.

\section{Ruling out memorization of the target papers}\label{app:leakage}
The cutoff-axis findings could be confounded if newer models simply memorized the specific papers behind each subfield. Three checks rule this out. (i)~Every retained source paper is published \emph{after} the knowledge cutoff of nearly all scored models, so by construction the target literature is unseen at training time. (ii)~A model-demeaned regression of score on the recency gap between a model's cutoff and each paper is flat, giving Static $+0.007$/yr at $p{=}0.75$ and Active $-0.012$/yr at $p{=}0.85$, both with $R^2\approx0$. (iii)~A direct probe asks each model to recite each target paper's background and reference list. Mean ROUGE-L against the true text is only $0.014$ and $91\%$ of models self-report the paper as unknown. For reference lists, ROUGE-L is $0.060$ with $79\%$ unknown, and only the oldest model, mistral-7b-v0.1, confabulates up to $0.17$. Across the $21$ models in this probe, a set that includes several proprietary models outside the main roster, prior-art coverage anti-correlates with Static score at Pearson $r{=}{-}0.32$ and Spearman $\rho{=}{-}0.32$. The more of a target a model can reproduce, the slightly lower it scores, so memorization confers no measurable advantage. What the cutoff axis tracks is general capability improving with model generation, not recall of the specific target papers.

\section{Tool-budget saturation}\label{app:budget}

\begin{figure*}[t]
\centering
\includegraphics[width=\textwidth]{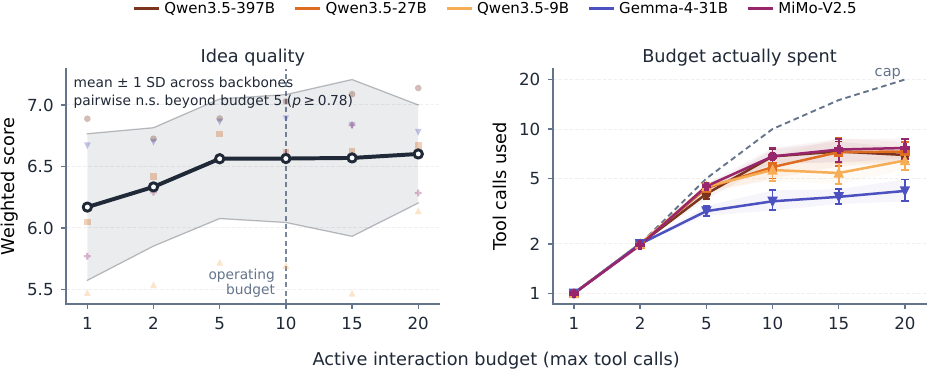}
\caption{Weighted score (left) and tool calls actually spent (right) against the Active tool-call budget, for five open-weight backbones collected in one batch. Each cell is $10$ source papers $\times$ $3$ hypotheses scored by $3$ static critics, aggregated critic~$\to$~idea~$\to$~paper~$\to$~cell. \textbf{Left:} the line is the mean over the five backbone cell means and the band is $\pm1$ standard deviation across those five, with each backbone's own cell mean drawn as a translucent marker; the dashed rule marks the operating budget of $10$. That mean is $6.56$, $6.56$, $6.57$, and $6.60$ at budgets $5$, $10$, $15$, and $20$, so it moves $+0.04$ from $10$ to $20$ against a between-backbone spread of $0.52$ at $10$; across the five backbone means no pair of budgets from $5$ upward differs significantly, $10$ versus $20$ included (paired $t$, $n{=}5$, all $p{\geq}0.78$). Paired per-paper tests find no backbone gaining between $5$ and $10$ calls, while one of the five still gains between $10$ and $20$. \textbf{Right:} per-backbone means with a $4000$-draw bootstrap $95\%$ interval over the $10$ papers, and the dashed diagonal marks the cap. All $30$ cells retain all $10$ papers.}
\label{fig:turnbudget}
\end{figure*}

We vary the Active tool-call budget over $\{1,2,5,10,15,20\}$ for five open-weight backbones: the qwen3.5 9b/27b/397b size ladder, plus gemma-4-31b and mimo-v2.5 (Figure~\ref{fig:turnbudget}). Every cell is $10$ papers $\times$ $3$ hypotheses $\times$ $3$ critics: $899$ rollouts of a planned $900$, one lost to a persistent search-API outage, and $2{,}643$ critic scores. All rows come from one collection batch, so budget is not confounded with collection date.

\paragraph{Saturation.} Every budget is run on the same $10$ papers, so we test each step with a paired per-paper bootstrap over those papers, $10{,}000$ draws. From $1$ to $5$ calls two backbones gain, qwen3.5-27b by $+0.72$ $[+0.25,+1.20]$ and mimo-v2.5 by $+0.81$ $[+0.39,+1.18]$, while the other three move by $|d|\le0.25$ at $p\ge0.27$. From $5$ to $10$ calls nothing moves at all: every backbone has $|d|\le0.15$ and $p\ge0.30$. From $10$ to $20$ one backbone gains again, qwen3.5-9b by $+0.45$ $[+0.20,+0.74]$, while the other four stay within $[-0.31,+0.11]$. Total spread across the six budgets runs from $0.22$ (gemma-4-31b) to $1.07$ (mimo-v2.5), and gemma-4-31b is the only backbone whose best cell is the operating budget of $10$. So the saturation is real but bounded: the interval over which every backbone is individually flat is $5$ to $10$, and a budget of $10$ understates one of the five backbones rather than sitting in a flat region for all of them.

\paragraph{Utilization does not explain it.} At a budget of $20$, realized utilization spans $21\%$ (gemma-4-31b, $4.2$ calls) to $38\%$ (mimo-v2.5, $7.7$ calls), with the remaining three between $32\%$ and $37\%$. No backbone comes close to the cap, and utilization does not line up with benefit. The one backbone that gains from $10$ to $20$ calls, qwen3.5-9b, is the second-lowest spender at $6.4$ of $20$, while the heaviest spender, mimo-v2.5, moves furthest the other way at $-0.31$. At the operating budget of $10$ utilization spans $36\%$ (gemma-4-31b) to $68\%$ (qwen3.5-397b and mimo-v2.5), so the cap binds for none of them. The intuitive reading, that scores flatten because models stop searching on their own, is not supported: how much of the budget a model spends and whether extra budget helps it are unrelated across these five.

\paragraph{Protocol failures.} Rollouts that emit three consecutive malformed commands are recorded as failures and go unscored. One backbone accounts for nearly all of them. qwen3.5-397b degrades with budget, from $0\%$ of rollouts at budgets $1$, $2$, and $5$ to $13\%$, $10\%$, and $24\%$ at $10$, $15$, and $20$, which is what motivates its exclusion from the \swm{} probe (Appendix~\ref{sec:improving}). mimo-v2.5 fails once, at budget $10$; gemma-4-31b, qwen3.5-9b, and qwen3.5-27b never fail. Because the failures scatter across papers rather than concentrating, every one of the $30$ cells still retains all $10$ papers.

\section{Reproducibility details}\label{app:repro}

\begin{table}[h]
\centering\small
\begin{tabular}{lr}
\toprule
Component & Count \\
\midrule
Disciplines & 5 \\
Subfields & 100 \\
Models evaluated (total) & 35 \\
\quad primary roster (either track) & 30 \\
\quad\quad with complete Static/Active pairs & 28 \\
\quad held-out closed-source (Gemini) & 5 \\
\quad with matched Static/Active outputs & 33 \\
Tracks per subfield & 2 (Static, Active) \\
Hypotheses per cell (scored subset) & 3 \\
Subfields in 3-per-cell scored subset & 40 \\
Critics per hypothesis & 3 (open-weight) \\
Scoring dimensions & 5 \\
Critic calls (scored subset) & $\sim$21{,}000 \\
Knowledge-cutoff span & 2023--2026 \\
\bottomrule
\end{tabular}
\caption{\name{} scale. The 3-hypotheses-per-cell scored subset backs all reported statistics; the full 100-subfield set is generated in both tracks for coverage.}
\label{tab:scale}
\end{table}

\paragraph{Sampling and runtime.} Generation uses temperature $0.7$ and seed $42$, scoring temperature $0.0$ and seed $42$. The Active agent has \texttt{max\_tokens}$=6000$ and a budget of $10$ SEARCH and FETCH turns. API calls retry with exponential backoff over up to 6 attempts, base $1$s and cap $60$s, with rate-limit backoff at base $2$s and cap $120$s. All generation and scoring were run 2026-06-30 to 2026-07-12 via OpenRouter, with the SWM extensions running to 2026-07-21, and the Semantic Scholar snapshot is contemporaneous. Idea generation and scoring are idempotent and resume-safe.

\paragraph{Critic and aggregation.} Three open-weight critics score every hypothesis, namely \texttt{qwen/qwen3.6-plus}, \texttt{moonshotai/kimi-k2.6}, and \texttt{z-ai/glm-5.1}. Per dimension we take a \emph{trimmed mean} that drops the single highest of the three critic scores and averages the rest, which at $n{=}3$ is the mean of the lower two, then weight $O{:}2,I{:}1.5,F{:}1,C{:}0.5,S{:}0.5$ and normalize by $5.5$. Before scoring, three keyword queries are extracted from each hypothesis and up to eight prior-art papers are retrieved from Semantic Scholar, filtered to \emph{publications on or before 2026-05-31}. That is a single global idea-conception date applied to all model ideas rather than each model's individual cutoff, a conservative choice because it filters strictly more aggressively for older models. The retrieved evidence, holding paper IDs, titles, years, abstracts, and queries, is frozen in table \texttt{e13\_evidence}, of which $5{,}705$ rows back the open-weight main analysis out of $11{,}180$ total that also span the Gemini held-out set and the appendix experiments. The curated Static references are frozen in \texttt{subdomain\_refs} as paper IDs and abstracts, so both the critic's prior art and the Static reference sets are reproducible. Parsing failures are $233/20{,}826$ of critic calls, or $1.1\%$, and failed calls are retried, never silently dropped.

\paragraph{Subfield selection.} A deterministic rule fixes the 40 scored subfields. Within each discipline the 20 subfields are sorted alphabetically, and indices $\{0,5,10,15\}$ from the pilot and $\{2,7,12,17\}$ from the extension are taken, giving eight evenly-spaced subfields per discipline and $40$ in total. No seed or content-based selection is involved.

\paragraph{Exclusions.} An Active idea shorter than 30 words or with an empty \texttt{FINAL} counts as a protocol failure and goes unscored, leaving that model with fewer Active cells. Two reasoning-only models, \texttt{qwen/qwen3-235b-a22b-thinking-2507} and \texttt{qwen/qwen3-vl-8b-thinking}, reject the Static path and run Active-only, so the paired analyses use $28$ models. Four models lack a documented knowledge cutoff, namely \texttt{minimax/minimax-m2.7}, \texttt{qwen/qwen3-coder}, \texttt{xiaomi/mimo-v2.5-pro}, and \texttt{z-ai/glm-4.6}, and are dropped from the cutoff regressions alone, which leaves $n{=}24$ Static and $26$ Active in the cutoff-axis analyses. The 397b MoE is excluded from the \swm{} probe for Active-protocol fragility (Appendix~\ref{app:budget}).

\paragraph{Cutoff and parameter provenance.} Knowledge-cutoff dates are compiled from model cards and provider disclosures where available, taking the mid-month day when only a month is given. Several recent models' cutoffs are best-effort estimates and a few are undisclosed, which drops them as described above. Parameter counts in Table~\ref{tab:models} are total parameters in billions from official figures where disclosed, and MoE models list total rather than active parameters. Undisclosed counts are left blank rather than estimated, and models whose provider has not disclosed a count, such as Mistral-medium, are omitted from parameter-controlled analyses. These values feed only the $\log$-parameter partial correlation in the two-driver analysis of \S\ref{app:stats}, and no headline finding depends on them.

\begin{table*}[t]
\centering\footnotesize\setlength{\tabcolsep}{3.4pt}
\begin{tabular}{llrrc@{\hskip 1.1em}llrrc}
\toprule
Model (OpenRouter id) & Family & P(B) & Cutoff & Tr & Model (OpenRouter id) & Family & P(B) & Cutoff & Tr \\
\midrule
deepseek-r1-0528            & deepseek   & 671  & 2024.5 & S$+$A & qwen3-235b-a22b-thinking   & qwen       & 235$^\ast$ & 2025.2 & A\phantom{S$+$} \\
deepseek-v4-flash           & deepseek   & --   & 2026.0 & S$+$A & qwen3-30b-a3b-instruct     & qwen       & 30$^\ast$  & 2025.3 & S$+$A \\
deepseek-v4-pro             & deepseek   & 671  & 2026.0 & S$+$A & qwen3-32b                  & qwen       & 32   & 2024.8 & S$+$A \\
gemma-2-27b-it              & google     & 27   & 2024.5 & S$+$A & qwen3-8b                   & qwen       & 8    & 2024.8 & S$+$A \\
gemma-3-27b-it              & google     & 27   & 2024.6 & S$+$A & qwen3-coder                & qwen       & --   & --     & S$+$A \\
gemma-4-31b-it              & google     & 31   & 2025.0 & S$+$A & qwen3-vl-8b-thinking       & qwen       & 8    & 2025.2 & A\phantom{S$+$} \\
llama-3.1-8b-instruct       & meta-llama & 8    & 2023.9 & S$+$A & qwen3.5-27b                & qwen       & 27   & 2025.3 & S$+$A \\
llama-4-maverick            & meta-llama & 400$^\ast$ & 2024.6 & S$+$A & qwen3.5-397b-a17b          & qwen       & 397$^\ast$ & 2025.3 & S$+$A \\
minimax-m2.7                & minimax    & --   & --     & S$+$A & qwen3.5-9b                 & qwen       & 9    & 2025.3 & S$+$A \\
mistral-medium-3.1          & mistralai  & --   & 2025.4 & S$+$A & mimo-v2.5                  & xiaomi     & 7    & 2025.4 & S$+$A \\
mistral-small-24b-2501      & mistralai  & 24   & 2023.8 & S$+$A & mimo-v2.5-pro              & xiaomi     & --   & --     & S$+$A \\
mistral-small-2603          & mistralai  & 24   & 2025.5 & S$+$A & glm-4.5-air                & z-ai       & 12   & 2025.2 & S$+$A \\
kimi-k2.5                   & moonshotai & 1000$^\ast$ & 2025.0 & S$+$A & glm-4.6               & z-ai       & --   & --     & S$+$A \\
kimi-k2.6                   & moonshotai & 1000$^\ast$ & 2025.0 & S$+$A & glm-5.1              & z-ai       & 110  & 2025.9 & S$+$A \\
qwen-2.5-72b-instruct       & qwen       & 72   & 2024.5 & S$+$A & qwen-2.5-7b-instruct       & qwen       & 7    & 2024.5 & S$+$A \\
\bottomrule
\end{tabular}
\caption{The 30 idea-generating models. \textbf{P(B)} is total parameters in billions, where $\ast$ marks an MoE total rather than an active count and some values are best-effort estimates, and \texttt{--} marks an undisclosed count. \textbf{Cutoff} is the knowledge cutoff as a decimal year, with \texttt{--} marking an undisclosed date that is dropped from the cutoff regressions. \textbf{Tr} lists the tracks generated, where S$+$A is both Static and Active and A is an Active-only reasoning model. Critic--generator overlap and self-grading robustness are in Appendix~\ref{app:judge}.}
\label{tab:models}
\end{table*}

\section{Artifact documentation and licensing}\label{app:license}
This section states the actual terms under which each input artifact was used and each released artifact is distributed.

\paragraph{Literature metadata and abstracts (Semantic Scholar).} Paper metadata, abstracts, and reference lists were obtained through the Semantic Scholar Academic Graph API under the Semantic Scholar API License Agreement. Academic Graph metadata is distributed under ODC-BY~1.0, which permits reuse with attribution. Individual records may carry other licenses such as CC BY-NC, and abstracts originating from publishers remain third-party content that the API agreement does not license us to redistribute. We give the required attribution to Semantic Scholar and cite the Open Data Platform.

\paragraph{Abstract redistribution.} We republish no abstracts. Abstracts and reference lists were used only as model input and as critic evidence at run time. Our released tables carry Semantic Scholar \texttt{corpusId} and DOI, retrieval timestamps, and our derived fields (query strings, relevance ranks, per-dimension scores). They carry no abstract text, no reference-list text, and no full text, so anyone reproducing the benchmark refetches that content from the API under their own agreement.

\paragraph{Generated hypotheses.} OpenRouter's terms defer output rights to each model's own terms, so redistribution rights follow the model. For the open-weight roster the weight licenses in Table~\ref{tab:licenses} permit us to release the generated hypotheses, and we do. The five held-out Gemini models were accessed through the Gemini API, whose additional terms assign ownership of generated output to the developer while forbidding the use of that output to build or improve a competing model. We therefore withhold their generated text entirely, and for those five models the release carries scores, per-dimension breakdowns, and derived statistics only. Releasing the text under our data license would purport to grant rights we do not hold.

\paragraph{Upstream model licenses.} Table~\ref{tab:licenses} lists the license each family states on its model card. Two deserve naming rather than folding into ``permissive''. The Gemma Terms of Use and the Llama Community License are custom licenses with acceptable-use policies and downstream flow-down obligations, and neither is OSI-approved, while the MiniMax license permits non-commercial use and requires written authorization for commercial use. We therefore describe the roster as \emph{open-weight} rather than open-source throughout the paper. One roster entry runs the other way. \texttt{mistral-medium-3.1} is a premier API model with no public weights, and it stays in the roster because it is scored under exactly the same protocol as the rest, not because it is redistributable.

\paragraph{Other upstream artifacts.} Idea-diversity embeddings use \texttt{sentence-transformers} with \texttt{all-MiniLM-L6-v2}, both under Apache-2.0. Reference-ordering embeddings use OpenAI \texttt{text-embedding-3-small} through its API, and we release the resulting rank orders rather than the vectors. The pinned runtime dependencies \texttt{openai}, \texttt{requests}, and \texttt{python-dotenv}, together with the analysis stack \texttt{numpy}, \texttt{scipy}, and \texttt{matplotlib}, are all under Apache-2.0, BSD, or MIT terms.

\paragraph{Our release.} The benchmark code, the critic prompts, and the scoring pipeline are released under the MIT License. The generated hypotheses, critic scores, and derived statistics are released under CC BY 4.0, which matches the attribution obligation we inherit from ODC-BY upstream. Both licenses ship in the repository, \url{https://github.com/HKUST-KnowComp/AgentIdeaBench}.

\paragraph{Intended use.} The artifact supports research on evaluating and improving language-model ideation, covering reproduction of our numbers, addition of models to the roster, and study of critic behavior. It is not a validated instrument for deciding what science to fund or publish, and the hypotheses in it are untested model output rather than vetted scientific claims (see the Ethics Statement). We intend use consistent with the ACL Code of Ethics, and we do not sanction using it to mass-produce unreviewed research proposals.

\begin{table}[t]
\centering\footnotesize\setlength{\tabcolsep}{5pt}\renewcommand{\arraystretch}{0.95}
\begin{tabular}{ll}
\toprule
Family (models used) & Weights license \\
\midrule
qwen (incl.\ coder, VL)    & Apache-2.0 \\
deepseek (R1, V4)          & MIT \\
z-ai (GLM)                 & MIT \\
xiaomi (MiMo)              & MIT \\
moonshotai (Kimi)          & Modified MIT \\
minimax (M2.7)             & Modified MIT, non-commercial \\
mistralai (Small)          & Apache-2.0 \\
mistralai (Medium 3.1)     & Premier, API only \\
google (Gemma)             & Gemma Terms of Use \\
meta-llama (3.1, 4)        & Llama Community License \\
\midrule
google (Gemini, held out)  & API only, no weights \\
\bottomrule
\end{tabular}
\caption{License each model family states for its released weights, as of the model cards at the time of writing. The held-out Gemini models are API-only; their outputs are governed by the Gemini API terms rather than a weights license.}
\label{tab:licenses}
\end{table}

\section{Statistical reporting}\label{app:stats}
We summarize the unit of analysis, test, effect size, and uncertainty for each headline claim; all bootstraps use $5000$ resamples unless noted, seed fixed.

\paragraph{Capability gate.} The unit of analysis is the model, with $n{=}28$ carrying both tracks. Pearson $r(\text{Static},\text{gain}){=}0.69$ at $p{<}10^{-4}$, and Spearman $\rho{=}0.63$. The correlation strengthens monotonically with hypotheses sampled per cell, with Pearson $r$ moving $0.56{\to}0.64{\to}0.69$ at $1{\to}2{\to}3$ hypotheses per cell for $n{=}28$, which is the opposite of what a small-sample artifact would do. For robustness to the family-heavy roster, where 9 of 28 models are qwen, a cluster bootstrap over the nine families gives 95\% CI $[0.56,0.85]$, and leave-one-family-out keeps $r\in[0.67,0.76]$. \emph{Split-half.} Static appears on both axes, since the gain $C{-}B$ shares $B$'s noise with the abscissa. We therefore recompute the gate with ability measured on one random half of the $40$ subfields and the gain on the disjoint other half, so the two carry independent scoring noise. Over $2000$ random splits Pearson $r$ averages $0.70$, with 95\% CI $[0.64,0.75]$ and a positive value in every split. The knowledge-cutoff axis, which shares no scoring noise with the gain, predicts the gain independently at $r{=}0.69$, $p{=}0.0002$, and $n{=}24$. \emph{Held-out closed frontier family.} We additionally score five Gemini models, namely \texttt{gemini-2.5-flash-lite}, \texttt{-2.5-flash}, \texttt{-3-flash-preview}, \texttt{-3.1-pro-preview}, and \texttt{-3.5-flash}, on the identical 40 subfields and pipeline. We exclude \texttt{gemini-3.6-flash} for leakage, because its cutoff post-dates the benchmark. None appears in the primary roster, which makes this a true out-of-roster test. The gate replicates. The weakest model, at static $4.63$, is hurt by $-0.37$, the next is flat at $-0.05$, and the three strongest gain between $+0.35$ and $+0.70$. Pearson $r{=}0.88$ across the five at $p{=}0.05$ and $n{=}5$, a sample small enough that we read the ordering rather than the interval. Pooling all $33$ models keeps a strong positive gate at $r{=}0.57$, $p{<}10^{-3}$, and 95\% CI $[0.34,0.74]$.

\paragraph{Discrimination.} The unit of analysis is the model, with $n{=}28$, bootstrapped over the 40 subfields across $5000$ resamples for per-model standard errors. Across the roster the Active total has $4.4\times$ the between-model variance of the Static total (95\% CI $[3.5,5.2]$). The eight models Static ranks highest span $0.39$ under Static and $1.39$ under Active (paired bootstrap difference $+0.96$, 95\% CI $[0.64,1.26]$, one-sided $p{<}10^{-3}$). Treating two models as distinguishable when their means differ by more than twice the pooled bootstrap SE, Active separates $87\%$ of all $378$ model pairs and $62\%$ of pairs within the top half of the roster, against Static's $60\%$ and $12\%$. The top-end gain is a real increase in resolving power, measured over all pairs rather than a hand-picked range.

\paragraph{Track$\times$cutoff scaling.} We fit \emph{score $\sim$ cutoff $+$ track $+$ cutoff$\times$track} directly instead of comparing two slopes. At model level, with $n{=}26$ models carrying cutoffs over $50$ model$\times$track rows, the interaction is $+0.62$/yr with a cluster-bootstrap-over-models 95\% CI $[0.41,0.94]$ and within-model track-permutation $p{<}10^{-3}$. At cell level, over $1996$ model$\times$subdomain$\times$track cells, the interaction is $+0.62$/yr with a two-way model$\times$subdomain cluster-bootstrap 95\% CI $[0.42,0.91]$. The descriptive per-track slopes are Static $+0.54$/yr and Active $+1.16$/yr. Fit separately within each discipline (Figure~\ref{fig:domainslopes}), the Static and Active slopes are chemistry $+0.58$ and $+1.21$, medicine $+0.41$ and $+1.19$, computer science $+0.70$ and $+1.18$, biology $+0.59$ and $+1.18$, and physics $+0.44$ and $+1.05$. Active exceeds Static in every discipline, by $+0.48$ to $+0.78$ points per year.

\paragraph{Two independent drivers, size and recency.} Ideation ability decomposes into more than one latent factor. Partial correlations on Static scores, over the $n{=}22$ models with disclosed size and cutoff, show that knowledge recency given size, at $+0.74$, and model size given recency, at $+0.64$, each independently predict quality, both at $p{<}0.005$. A small recent model and a large older model reach comparable Static quality by different routes, which is why we read ability against both the capability and cutoff axes throughout \S\ref{sec:results}.

\paragraph{Idea diversity.} At cell level, over 2793 model$\times$subdomain pairs, the Vendi gain is $+0.149$, but cells are correlated. Cluster-level tests across models, with $n{=}28$, give mean $+0.15$, Wilcoxon $p{=}0.004$, matched-pairs rank-biserial $0.60$, 20/28 positive, and 95\% CI $[0.07,0.23]$. Across subfields, with $n{=}100$, the gain is $+0.149$ with 95\% CI $[0.13,0.17]$. Capability against Vendi gain gives Pearson $r{=}{-}0.47$ at $p{=}0.011$ and $n{=}28$.

\paragraph{\swm{} (S4b).} The unit of analysis is the backbone$\times$subdomain cell. Against baseline, pooled over the four SWM backbones with $n{=}173$, the effect is $+0.245$ at Wilcoxon $p{=}0.042$ and 95\% CI $[+0.03,+0.46]$. Holm-corrected over the three designs \{S2,S4,S4b\} it gives $p_{\text{Holm}}{=}0.125$, with S4 at $0.15$ and S2 at $0.15$, leaving none significant. The direct S4b$-$S4 comparison, paired on the $180$ shared cells at $45$ per backbone, gives $+0.03$ at Wilcoxon $p{=}0.75$, rank-biserial $0.03$, and 95\% CI $[-0.18,+0.23]$. Per backbone it is $+0.29$ on 9b at $p{=}0.17$, $+0.07$ on 27b at $p{=}0.80$, $-0.14$ on v4-flash at $p{=}0.50$, and $-0.11$ on v4-pro at $p{=}0.62$. The point estimates order S4b $>$ S4 $>$ S2 and S4b is best on both qwen backbones, but no pairwise or corrected test establishes S4b $>$ S4.

\section{Complete evaluated roster}\label{app:fullroster}
Table~\ref{tab:leaderboard-all} lists every paired model we have scored, adding to the paper's roster the closed OpenAI and Anthropic routes we reached through an internal gateway: five 2026 frontier models and a 25-model OpenAI/Anthropic release ladder. We report them for completeness and treat them as held-out throughout, so no headline statistic in the main text is computed on them and none is bolded. Two routes are absent, \texttt{gpt-5-chat} and \texttt{gpt-5.3-chat}, each being the non-reasoning serving configuration of an already-listed model and so a second route to one model rather than a second model. Coverage of the ladder is near-complete: $87$ of $18{,}000$ scoring cells, or $0.48\%$, are missing, $27$ of them because one Anthropic route's safety filter rejects three biomedical subfields at input, so the affected models' means are computed over $351$--$357$ cells in place of $360$.

The capability gate of \S\ref{sec:results} is the claim most sensitive to this choice, so we state its behavior on the wider set explicitly. Pearson $r(\text{Static},\text{gain})$ is $+0.695$ at $p{=}4.1{\times}10^{-5}$ on the 28 open-weight models, $+0.311$ at $p{=}0.018$ on the 58 models that add the five held-out Gemini models and the 25-model closed release ladder, and $+0.182$ at $p{=}0.15$ on all 63 once the five 2026 frontier models are added as well. The gate survives those 30 added held-out models and does not survive the frontier five, which is consistent with the ceiling reading in \S\ref{sec:results}. We lack the controlled evidence to separate a genuine breakdown from scorer headroom, so we leave the headline claim scoped to the open-weight roster.

\begin{table*}[p]
\centering
\caption{\textbf{Complete evaluated roster.} The same measurement as Table~\ref{tab:leaderboard}, extended to all 63 paired models we have scored and ranked by Active total. All closed-source models are held out, marked $\dagger$, and excluded from every headline statistic, which stays on the 28 open-weight models, so best-in-column \textbf{bold} still marks the best open-weight value. Columns, weighting, and critic are exactly as in Table~\ref{tab:leaderboard}. Roster composition and coverage notes are in the text above.}
\label{tab:leaderboard-all}
\scriptsize\setlength{\tabcolsep}{3pt}\renewcommand{\arraystretch}{1.10}
\begin{tabular*}{\textwidth}{@{\extracolsep{\fill}} l rrr r rrrrr @{}}
\toprule
& \multicolumn{3}{c}{Weighted total} & & \multicolumn{5}{c}{Active per-dimension} \\
\cmidrule(lr){2-4} \cmidrule(lr){6-10}
Model & Static & Active & $\Delta$ & Turns & Orig. & Feas. & Clar. & Impact & Spec. \\
\midrule
Claude Opus 5$^\dagger$ & 6.89 & 7.14 & $+0.25$ & 9.7 & 7.30 & 6.77 & 7.87 & 6.75 & 7.63 \\
GPT-5.6 Sol$^\dagger$ & 6.90 & 6.88 & $-0.02$ & 9.8 & 6.93 & 6.72 & 7.78 & 6.40 & 7.47 \\
GPT-5.6 Terra$^\dagger$ & 6.86 & 6.80 & $-0.06$ & 9.7 & 6.81 & 6.76 & 7.71 & 6.32 & 7.37 \\
Claude Opus 4.8$^\dagger$ & 5.78 & 6.79 & $+1.00$ & 7.0 & 6.81 & 6.83 & 7.60 & 6.30 & 7.24 \\
Claude Opus 4.7$^\dagger$ & 5.83 & 6.70 & $+0.87$ & 9.9 & 6.71 & 6.49 & 7.70 & 6.29 & 7.30 \\
GPT-5.5$^\dagger$ & 6.70 & 6.68 & $-0.01$ & 10.0 & 6.70 & 6.63 & 7.51 & 6.28 & 7.12 \\
Claude Sonnet 5$^\dagger$ & 6.05 & 6.66 & $+0.61$ & 7.1 & 6.70 & 6.52 & 7.59 & 6.22 & 7.21 \\
GPT-5.2$^\dagger$ & 6.16 & 6.66 & $+0.50$ & 8.0 & 6.58 & 7.00 & 7.60 & 6.06 & 7.11 \\
GPT-5.4$^\dagger$ & 6.18 & 6.62 & $+0.44$ & 8.7 & 6.51 & 7.00 & 7.39 & 6.15 & 6.99 \\
GPT-5.6 Luna$^\dagger$ & 6.70 & 6.43 & $-0.27$ & 9.7 & 6.28 & 6.63 & 7.41 & 6.00 & 6.91 \\
GPT-5$^\dagger$ & 6.55 & 6.40 & $-0.15$ & 9.8 & 6.29 & 6.46 & 7.51 & 5.91 & 7.06 \\
GPT-5 mini$^\dagger$ & 5.88 & 6.38 & $+0.49$ & 9.1 & 6.12 & 6.75 & 7.43 & 5.91 & 6.97 \\
GPT-5.4 mini$^\dagger$ & 6.06 & 6.36 & $+0.30$ & 5.1 & 6.20 & 6.85 & 7.17 & 5.87 & 6.67 \\
GLM-5.1 & 5.13 & \textbf{6.33} & $\mathbf{+1.21}$ & 8.8 & \textbf{6.31} & 6.32 & \textbf{7.37} & \textbf{5.90} & \textbf{6.71} \\
Gemini-3 Flash$^\dagger$ & 5.59 & 6.29 & $+0.70$ & 8.5 & 6.33 & 6.16 & 7.12 & 5.92 & 6.60 \\
Gemini-3.5 Flash$^\dagger$ & 5.87 & 6.22 & $+0.35$ & 9.9 & 6.09 & 6.51 & 7.21 & 5.77 & 6.56 \\
Gemini-3.1 Pro$^\dagger$ & 5.63 & 6.20 & $+0.57$ & 7.5 & 6.20 & 6.14 & 7.14 & 5.83 & 6.47 \\
Kimi-K2.6 & 5.10 & 6.17 & $+1.08$ & 9.8 & 6.08 & 6.38 & 7.22 & 5.70 & 6.53 \\
o3$^\dagger$ & 5.56 & 6.06 & $+0.49$ & 5.5 & 5.94 & 6.10 & 7.38 & 5.57 & 6.60 \\
GPT-5.1$^\dagger$ & 5.66 & 6.04 & $+0.38$ & 4.0 & 5.90 & 6.22 & 6.90 & 5.74 & 6.26 \\
Claude Opus 4.6$^\dagger$ & 5.80 & 6.03 & $+0.24$ & 10.0 & 5.68 & 6.70 & 7.22 & 5.52 & 6.50 \\
Qwen3.5 397B & 5.06 & 5.95 & $+0.89$ & 8.9 & 5.79 & 6.28 & 6.81 & 5.57 & 6.19 \\
Kimi-K2.5 & 4.91 & 5.92 & $+1.02$ & 9.9 & 5.84 & 6.00 & 6.86 & 5.58 & 6.17 \\
DeepSeek-V4 Pro & \textbf{5.32} & 5.92 & $+0.60$ & 8.5 & 5.79 & 5.94 & 7.14 & 5.53 & 6.35 \\
MiMo-V2.5 Pro & 5.11 & 5.91 & $+0.81$ & 9.0 & 5.63 & 6.46 & 6.97 & 5.46 & 6.26 \\
Claude Sonnet 4.6$^\dagger$ & 5.36 & 5.84 & $+0.48$ & 10.0 & 5.56 & 6.10 & 7.09 & 5.46 & 6.31 \\
GPT-5.4 nano$^\dagger$ & 5.72 & 5.82 & $+0.10$ & 4.6 & 5.50 & 6.64 & 6.77 & 5.29 & 6.12 \\
GLM-4.6 & 4.83 & 5.78 & $+0.95$ & 6.9 & 5.64 & 6.14 & 6.84 & 5.31 & 5.99 \\
Claude Opus 4.5$^\dagger$ & 5.52 & 5.62 & $+0.10$ & 10.0 & 5.18 & 6.48 & 6.74 & 5.12 & 6.00 \\
o4-mini$^\dagger$ & 5.39 & 5.61 & $+0.22$ & 7.6 & 5.31 & 6.12 & 6.97 & 5.05 & 6.18 \\
Qwen3.5 27B & 4.71 & 5.53 & $+0.82$ & 8.0 & 5.27 & 5.82 & 6.64 & 5.22 & 5.83 \\
DeepSeek-V4 Flash & 5.09 & 5.52 & $+0.42$ & 9.0 & 5.14 & 6.16 & 6.70 & 5.09 & 5.82 \\
Claude Sonnet 4.5$^\dagger$ & 5.02 & 5.50 & $+0.48$ & 10.0 & 5.28 & 5.83 & 6.50 & 5.18 & 5.67 \\
Gemma-4 31B & 5.14 & 5.49 & $+0.35$ & 5.0 & 5.22 & 6.05 & 6.78 & 4.97 & 5.75 \\
MiMo-V2.5 & 4.90 & 5.46 & $+0.56$ & 7.8 & 5.14 & 6.05 & 6.35 & 5.17 & 5.55 \\
Claude Haiku 4.5$^\dagger$ & 5.20 & 5.37 & $+0.17$ & 9.9 & 5.11 & 5.89 & 6.32 & 5.04 & 5.47 \\
Mistral Medium 3.1 & 4.74 & 5.37 & $+0.62$ & 6.7 & 5.00 & 5.82 & 6.52 & 5.08 & 5.63 \\
GPT-5 nano$^\dagger$ & 5.75 & 5.36 & $-0.39$ & 6.7 & 4.83 & 6.38 & 6.51 & 4.92 & 5.62 \\
MiniMax-M2.7 & 4.70 & 5.33 & $+0.62$ & 9.7 & 4.95 & 6.03 & 6.47 & 4.91 & 5.50 \\
DeepSeek-R1 & 4.86 & 5.30 & $+0.44$ & 6.0 & 5.08 & 5.50 & 6.49 & 4.95 & 5.68 \\
Mistral Small 2603 & 4.61 & 5.16 & $+0.55$ & 9.4 & 4.65 & 5.93 & 6.60 & 4.72 & 5.55 \\
Gemini-2.5 Flash$^\dagger$ & 5.10 & 5.05 & $-0.05$ & 7.7 & 4.53 & 6.07 & 6.01 & 4.78 & 4.92 \\
Qwen3 30B & 4.93 & 4.94 & $+0.01$ & 8.3 & 4.49 & 5.65 & 6.17 & 4.58 & 5.18 \\
Qwen3.5 9B & 4.23 & 4.87 & $+0.64$ & 6.9 & 4.75 & 4.90 & 6.03 & 4.53 & 5.13 \\
Gemma-3 27B & 4.68 & 4.82 & $+0.14$ & 6.2 & 4.16 & 6.34 & 6.09 & 4.21 & 5.02 \\
GLM-4.5 Air & 4.31 & 4.82 & $+0.51$ & 8.9 & 4.49 & 5.60 & 5.56 & 4.59 & 4.56 \\
o3-mini$^\dagger$ & 4.84 & 4.82 & $-0.02$ & 3.6 & 4.27 & 6.20 & 5.55 & 4.49 & 4.54 \\
Qwen3 Coder & 4.87 & 4.72 & $-0.15$ & 8.6 & 3.82 & 6.55 & 6.15 & 4.14 & 5.02 \\
GPT-4.1$^\dagger$ & 4.92 & 4.70 & $-0.22$ & 8.7 & 3.80 & 6.62 & 5.88 & 4.26 & 4.65 \\
o1$^\dagger$ & 4.55 & 4.66 & $+0.11$ & 8.7 & 4.03 & 6.03 & 5.74 & 4.23 & 4.66 \\
Qwen3 32B & 4.53 & 4.44 & $-0.10$ & 3.1 & 4.13 & 5.04 & 5.31 & 4.15 & 4.46 \\
GPT-4o$^\dagger$ & 4.71 & 4.37 & $-0.34$ & 9.9 & 3.40 & 6.62 & 5.47 & 3.88 & 4.15 \\
GPT-4.1 nano$^\dagger$ & 4.66 & 4.32 & $-0.35$ & 2.0 & 3.43 & 6.38 & 5.20 & 3.95 & 3.94 \\
Gemini-2.5 Flash-Lite$^\dagger$ & 4.63 & 4.27 & $-0.37$ & 6.9 & 3.20 & 6.71 & 5.47 & 3.70 & 4.15 \\
GPT-4.1 mini$^\dagger$ & 4.92 & 4.26 & $-0.66$ & 7.3 & 3.12 & 6.62 & 5.52 & 3.78 & 4.26 \\
Qwen3 8B & 4.24 & 4.21 & $-0.03$ & 4.3 & 3.84 & 5.10 & 4.95 & 3.90 & 4.13 \\
Qwen2.5 72B & 4.20 & 3.97 & $-0.23$ & 8.8 & 2.67 & 6.77 & 5.33 & 3.37 & 3.96 \\
Mistral Small 24B & 4.19 & 3.93 & $-0.26$ & 8.9 & 2.88 & 6.25 & 5.01 & 3.48 & 3.82 \\
Llama-4 Maverick & 4.34 & 3.84 & $-0.50$ & 4.5 & 2.63 & \textbf{6.87} & 4.77 & 3.23 & 3.47 \\
GPT-4o mini$^\dagger$ & 4.09 & 3.76 & $-0.33$ & 9.2 & 2.58 & 6.66 & 4.76 & 3.20 & 3.41 \\
Qwen2.5 7B & 3.97 & 3.67 & $-0.30$ & 8.8 & 2.62 & 6.12 & 4.45 & 3.28 & 3.31 \\
Llama-3.1 8B & 3.80 & 3.54 & $-0.26$ & 6.7 & 2.65 & 5.66 & 4.11 & 3.29 & 3.06 \\
Gemma-2 27B & 4.02 & 3.20 & $-0.82$ & 4.1 & 2.26 & 5.60 & 3.90 & 2.75 & 2.75 \\
\bottomrule
\end{tabular*}
\end{table*}

\section{Recall-only control (closed-book)}\label{app:recall}
The Static/Active contrast could be confounded by mere reference presence, if being handed any references at all were what drives the effect. To test that we run a third track, \textbf{recall-only}, in which the model is given only the discipline and subfield name, with neither the curated references of Static nor the search tools of Active, and must produce a hypothesis from parametric knowledge alone. We generate it for the $28$ both-track models on $10$ subfields, two per discipline, taken as the alphabetical-index-$\{0,5\}$ subset of the scored $40$. Each cell holds three hypotheses, scored with the identical three-critic lit8d pipeline over $2{,}520$ critic calls. Static and Active are re-aggregated on exactly these $28$ models and $10$ subfields so all three are compared on the same $280$ (model, subfield) cells.

\paragraph{Result.} Mean weighted scores are R $4.82$, B (Static) $4.72$, and C (Active) $5.05$. Paired Wilcoxon tests over the $280$ shared cells give the following.
\begin{itemize}\itemsep2pt
\item \textbf{Static$-$Recall} ${=}{-}0.09$ at win rate $0.45$, $p{=}0.11$, and 95\% CI $[-0.20,+0.01]$. This is not significant, so curated references do not raise hypothesis quality above closed-book recall.
\item \textbf{Active$-$Recall} ${=}{+}0.23$ at win rate $0.59$, $p{<}10^{-3}$, and 95\% CI $[+0.12,+0.34]$, so Active sits significantly above recall.
\item \textbf{Active$-$Static} ${=}{+}0.33$ at win rate $0.65$, $p{<}10^{-3}$, and 95\% CI $[+0.22,+0.43]$, consistent with the main Active$>$Static result.
\end{itemize}

\paragraph{Reading.} Recall-only does not sit as a clean floor below both retrieval regimes. The ordering is \textbf{Static$\approx$Recall$<$Active}. Handing a model curated references under Static yields no measurable gain over asking it to recall from parametric knowledge, whereas letting it control its own retrieval under Active does. That sharpens the construct rather than weakening it, since the operative variable is agent-controlled retrieval rather than reference presence as such. One caveat applies. Recall-only is measured on the $10$-subfield, $28$-model subset for exact pairing, and the sign of Static$-$Recall is negative but within noise, so the claim we make is indistinguishability, written Static$\approx$Recall, and never that references hurt.

\section{Replay control: content versus process}\label{app:replay}
\begin{figure}[t]
\centering
\includegraphics[width=\columnwidth]{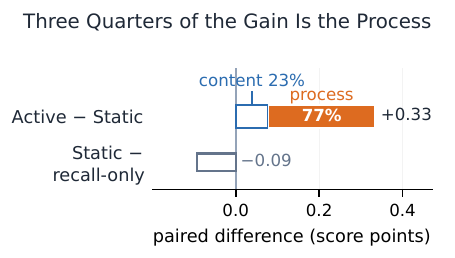}
\caption{Where the Active gain comes from, at cell level. The total stacks into a content term, replay $-$ Static, and a process term, Active $-$ replay, so each term's share is its length. The recall control below is a separate contrast rather than a slice of that total. Hollow markers are not significant. See \S\ref{sec:altrobust} and Appendix~\ref{app:recall}.}
\label{fig:decomp}
\end{figure}
The recall-only control (Appendix~\ref{app:recall}) shows that mere reference \emph{presence} does not drive the Active gain, since Static${\approx}$Recall${<}$Active and the model must gather its own evidence. This appendix adds the complementary arm, which splits gathering into the \emph{content} the agent surfaces and the agentic \emph{process} of surfacing it.

\paragraph{Design.} The \textbf{replay} track feeds each model, in the passive single-pass Static format (\S\ref{sec:setup-eval}), the exact papers its own Active agent retrieved, and asks for a hypothesis in the identical format. Those papers are reconstructed from the stored Active telemetry as deduplicated \texttt{search\_papers} hits carrying title and abstract, in encounter order, capped at the Static reference count of $10$. Only the reference source differs from Static, being agent-surfaced where Static is designer-curated, while the prompt machinery, output format, and reference count are held fixed. We run replay for the $28$ matched primary-roster models on the same $10$ subfields as the recall control, with three hypotheses per cell, score it with the identical three-critic lit8d pipeline, and re-aggregate Static and Active on the shared cells for exact pairing. Three contrasts then sit on a common footing.
\begin{itemize}\itemsep2pt
\item Replay$-$Static is the reference content effect, agent-surfaced against curated, with delivery held passive;
\item Active$-$Replay is the agentic process effect, multi-turn control against single-pass, with content held fixed;
\item Active$-$Static is the total Active gain, decomposed additively as (Replay$-$Static)$+$(Active$-$Replay).
\end{itemize}

\paragraph{Result.} Mean weighted scores on the $279$ shared cells are Replay $4.80$, Static $4.72$, and Active $5.06$. Paired Wilcoxon tests over the shared cells give the following.
\begin{itemize}\itemsep2pt
\item Replay${-}$Static${=}{+}0.08$ at win rate $0.52$, $p{=}0.21$, and 95\% CI $[-0.03,+0.19]$. This is not significant, so the reference content the agent surfaces, delivered passively, does not outscore the curated Static set.
\item Active${-}$Replay${=}{+}0.26$ at win rate $0.60$, $p{<}10^{-3}$, and 95\% CI $[+0.15,+0.37]$, so the agentic process of gathering the evidence across turns carries the gain.
\item Active${-}$Static${=}{+}0.33$ at win rate $0.66$, $p{<}10^{-3}$, and 95\% CI $[+0.23,+0.44]$, which is the total Active gain and matches both the main result and the recall control's $+0.33$.
\end{itemize}
The decomposition is stable at the model level, where $n{=}28$ gives Replay${-}$Static${=}{+}0.08$ at $p{=}0.19$, Active${-}$Replay${=}{+}0.25$ at $p{=}0.02$, and Active${-}$Static${=}{+}0.33$ at $p{=}0.002$.

\paragraph{Reading.} The realized pattern is Replay${\approx}$Static${<}$Active. The content term is near zero and not significant, and the process term carries the gain. The Active advantage thus lies in the agentic process of deciding what to read across turns rather than in the specific papers surfaced. Passively handing a model its own agent's references reproduces the Static score, not the Active one. Together with the recall control, where Static${\approx}$Recall${<}$Active (Appendix~\ref{app:recall}), this brackets the effect from both sides, since neither the presence of references nor their content, delivered passively, matches agent-controlled retrieval. This arm still leaves multi-turn interaction bundled with tool-execution competence inside the process term (\S\ref{sec:limits}).

\section{Generation-time improvement: scaffolds and a scientific-world-model probe}\label{sec:improving}

This appendix supplements the main-text \swm{} analysis (\S\ref{sec:swm-main}) with the design and control details it omits. We first establish an anti-Goodhart baseline. Single pre-generation scaffolds each move only one dimension and cannot shift the originality-weighted total (\S\ref{sec:scaffold}), which is what motivates a closed loop at all. We then give the \swm{}'s full interface and five internal designs (\S\ref{sec:swm}) and the gated S5 variant (\S\ref{sec:s5}). The compute-matched best-of-3 baseline is in Appendix~\ref{app:matched}, and the chain-of-thought internalization probe in Appendix~\ref{app:cot}.

\subsection{Single scaffolds fail (anti-Goodhart)}\label{sec:scaffold}
We test two pre-generation scaffolds in Static mode, meaning prompts that restructure the model's thinking before it commits. A \textbf{grounding scaffold} asks the model to first write the field's known mechanisms and open problems, then propose. It raises feasibility by $+0.32$ and leaves originality alone, so the weighted total barely moves at $+0.08$. A \textbf{cross-domain analogy} scaffold asks the model to import a concept from a distant field. It raises originality by $+0.30$ and loses feasibility at $-0.11$, again leaving the total flat at $+0.08$, and the effect is itself capability-gated, reaching $+1.07$ on the largest model while going negative on small ones. Each scaffold pulls a single lever at the other's expense. For the benchmark this is a desirable property, since an originality-weighted total that a one-shot prompt cannot move is \textbf{resistant to cheap Goodhart-style gaming}. It also sets the bar for a real method, which has to move originality and feasibility together within a single loop where these scaffolds trade one for the other.

\subsection{Closed-loop Scientific World Model}\label{sec:swm}
We extend the Active protocol with a fourth command. \texttt{SIMULATE:\,<draft>} sends the current draft to a black-box \swm{}, which returns structured feedback containing a mechanism-consistency verdict; a thought-experiment prediction, explicitly flagged as a plausibility signal and not truth; the \textbf{novel core}, one sentence that must survive any edit; per-dimension weaknesses with novelty and feasibility on separate channels; a \textbf{feasibility repair} constrained not to weaken the novel core; a knowledge gap; and a suggested next action among search, refine, and accept. The agent must simulate at least once before finalizing. Two design commitments define the interface. The first decouples novelty from feasibility onto separate channels, and the second pins a protected core that the feasibility repair may not touch. That is the interface a single scaffold lacks, and we conjecture it is the mechanism that lets one loop raise both levers at once. The experiments below probe that conjecture without confirming it at this sample size.

\paragraph{Relation to active inference.} The interface lines up with the expected-free-energy decomposition at three points, and that correspondence is what motivated its design (\S\ref{sec:swm-main}). The two channels the schema holds apart, novelty and feasibility, stand in for the epistemic and pragmatic terms, and the single-scaffold results of \S\ref{sec:scaffold} show the two trading against each other whenever a one-shot prompt pushes only one. The protected \texttt{novel\_core} exists because an unconstrained feasibility repair spends the epistemic term to buy the pragmatic one, and the per-dimension diagnosis behind S5 (\S\ref{sec:s5}) is that trade showing up in the scores. \texttt{SIMULATE} is a counterfactual rollout under the agent's own generative model, in the sense planning-as-inference gives to policy evaluation \citep{friston2017processtheory,parr2022activeinference}. The capability gate of \S\ref{sec:gate} follows the same logic, since the worth of an epistemic action depends on the generative model that evaluates it, so a backbone whose model is poor gains little either from sampling the literature or from simulating its own draft. This is conceptual framing and nothing more. We optimize no variational objective, estimate no free energy, and report no result that tests the account.

\paragraph{Five internal designs.} The \swm{} is a black box, so we ablate what should be inside it while holding the agent-facing interface fixed. \textbf{S1} is a single inference in which one prompt plays all roles. \textbf{S2} is a fixed multi-role design holding a Simulator, a Novelty-scout that names and pins the novel core, a Feasibility-critic whose repair must not weaken it, and a deterministic referee. \textbf{S3} is S2 $+$ private retrieval grounding. \textbf{S4} is a dynamic meta-agent in which a LEAD convenes 2--4 bespoke experts per hypothesis, each with a role, focus, and optional retrieval tool, and then free-text aggregates their reviews. \textbf{S4b} pairs S4's dynamic panel with S2's hard-constrained assembly, so the panel's reviews become evidence for the fixed Novelty-scout $\rightarrow$ Feasibility-critic $\rightarrow$ referee chain. S1 and S2 vary the number of roles at fixed structure, while S4 and S4b share an identical dynamic panel and differ only in how the reviews are aggregated. Full schema and prompts are in Appendix~\ref{app:prompts}.

Per-backbone absolute scores are in the main text (\S\ref{sec:swm-main}, Table~\ref{tab:swm}), and the multiple-comparison correction, the direct S4b$-$S4 contrast, and the compute-matched baseline are in Appendices~\ref{app:stats} and~\ref{app:matched}--\ref{app:cot}. Two design details remain for this section, the S4-vs-S4b aggregation contrast and the gated S5 variant. The direct S4-vs-S4b test holds the dynamic panels identical and varies only the aggregator, and it comes out null, pooled $+0.03$ at $p{=}0.75$, with $+0.29$ and $+0.07$ on qwen against $-0.14$ and $-0.11$ on deepseek, so we make no claim that hard-constrained aggregation beats free-form aggregation. The pooled point estimates order S4b${>}$S4${>}$S2 at $+0.25$, $+0.21$, and $+0.17$, with the single-inference S1 at $+0.19$ and the retrieval-grounded S3 at $+0.32$, the latter on a smaller two-backbone sample, in the same fixed tier.

\paragraph{A gated repair does not recover the saturated family.}\label{sec:s5}
The per-dimension diagnosis suggests a candidate remedy. S4b's Feasibility-critic \emph{always} emits a repair, which on an already-good draft is a lateral edit that can trade away originality. We tested \textbf{S5} on the two deepseek backbones where the trade-off appears. S5 is S4b with a gate that suppresses the repair when feasibility is already strong, at severity $<2$, passing the draft through unedited. A small pilot at $n{=}31$/$17$ looked promising, and the full-sample run at $n{=}45$ per deepseek backbone \emph{did not confirm it}. S5 beats S4b on neither deepseek backbone, with matched S5$-$S4b at $-0.04$ and $-0.16$, and originality stays negative at $-0.42$ on v4-pro. The deepseek saturation is \textbf{robust to this intervention}, and no \swm{} variant we tried across S1--S5 lifts a backbone already at the score ceiling. Gating is sound in principle and does not help here, because the binding limit is the model's remaining originality headroom rather than the loop's editing discipline. We report this as a cautionary methodological case. The $n{\le}31$ pilot's positive sign reversed under the full sample, which is a reminder that directional conclusions from small ideation pilots are unreliable (Appendix~\ref{app:prompts}).

\section{Compute-matched baseline for \swm{}}\label{app:matched}
The closed-loop \swm{} spends more inference-time compute than plain Active. A single Active rollout on this probe's cells averages ${\sim}7$ LLM calls, namely $5.8$ tool turns plus a final synthesis, against a full-roster mean of $7.6$ over the 40 scored subfields (Table~\ref{tab:leaderboard}). S4b averages ${\sim}21$, being $8.4$ outer turns plus an internal expert panel of ${\sim}13$ calls, roughly $3\times$ as many. To separate S4b's aggregation structure from that extra compute, we compare it against a \textbf{best-of-3} baseline. The baseline runs three independent plain-Active rollouts per cell, ${\sim}21$ calls in total and matched to S4b, each scored by the identical lit8d pipeline, with the highest-scoring of the three taken as the cell output. We evaluate on the $173$ backbone-and-subfield cells for which all three rollouts and the S4b run are present, covering $4$ backbones at ${\sim}45$ cells each.

\paragraph{Result.} Mean weighted scores are single Active $5.31$, best-of-3 $6.10$, and S4b $5.55$. Paired Wilcoxon tests over the $173$ shared cells give the following.
\begin{itemize}\itemsep2pt
\item \textbf{best-of-3 $-$ single} ${=}{+}0.79$ at win rate $0.64$, $p{<}10^{-3}$, and 95\% CI $[+0.66,+0.93]$. Three independent samples plus selection lift the score far more than the loop does, and the effect is positive on all four backbones, ranging $+0.65$ to $+0.95$.
\item \textbf{S4b $-$ best-of-3} ${=}{-}0.55$ at win rate $0.35$, $p{<}10^{-3}$, and 95\% CI $[-0.73,-0.37]$. At equal compute S4b sits below the naive baseline, significantly so on three of four backbones, with v4-flash $-0.65$, v4-pro $-0.81$, and qwen-27b $-0.61$, all at $p{<}0.01$, and statistically tied on qwen-9b at $-0.05$ and $p{=}0.92$.
\item \textbf{S4b $-$ single} ${=}{+}0.245$ at win rate $0.57$, $p{=}0.042$, and 95\% CI $[+0.03,+0.46]$, which is the nominal S4b gain reproduced on these cells.
\end{itemize}

\paragraph{Reading.} S4b's nominal $+0.245$ over a single rollout is small beside the $+0.79$ that the \emph{same} compute yields under naive resampling, and S4b loses to the compute-matched best-of-3 on three of four backbones and ties on the fourth. We therefore cannot attribute the \swm{} gain to its aggregation structure, since at matched compute independent resampling matches or exceeds it everywhere. \textbf{Caveat.} Best-of-3 selects the highest of three drafts by the same critic score we then report. That oracle selection upper-bounds what extra compute buys and is no deployable method in itself, since the critic is unavailable at inference, which makes the claim analytical rather than a deployment recommendation. The loop leaves on the table value that even naive resampling captures, so the modest S4b effect is consistent with an inefficient use of inference-time compute and falls short of evidence for the aggregation mechanism. This is the strongest single reason we frame \swm{} as a promising but unconfirmed direction (Appendix~\ref{sec:improving}).

\section{Whether frontier backbones internalize the world-model check}\label{app:cot}
The closed-loop \swm{} helps the mid-capability qwen backbones and leaves the frontier deepseek-v4 backbones unchanged (Table~\ref{tab:swm}). Beyond the score-ceiling account (\S\ref{sec:swm-main}), a second candidate explanation is that a frontier model's own chain-of-thought already performs the novelty and mechanism checks the external \swm{} supplies, leaving little marginal room. We probe this directly and report it as an \emph{illustrative} observation rather than a confirmatory result, for the reasons stated at the end.

\paragraph{Setup.} The main pipeline stored no reasoning text, because generation ran with model ``thinking'' disabled for a clean cross-year comparison (Appendix~\ref{app:repro}). Neither \texttt{subdomain\_ideas} nor \texttt{swm\_ideas} retains any chain-of-thought, and the reasoning-token count is zero for every stored row. We therefore re-ran a small ideation set with thinking enabled and captured the chain-of-thought across eight subfields for the strongest backbone, deepseek-v4-pro, and for the backbone that gained most from \swm{}, qwen3.5-9b.

\paragraph{What the frontier backbone does unprompted.} In every captured trace, deepseek-v4-pro spontaneously audits its own idea against prior work and runs a compact thought experiment before committing. On a chronic-kidney-disease prompt it notes that ``MAA adducts are known in other diseases (alcohol liver disease, diabetes), but linking them to ficolin-3 and CKD progression might be novel,'' and on a CO$_2$-electrocatalysis prompt it asks ``is it novel? Cation effects are studied, but direct detection of cation redox state change is novel,'' then iterates to a less-explored variant. On a superconductivity prompt it derives a falsifiable consequence of its own proposed mechanism: the pairing gap ``scales inversely with the orbital splitting, and collapses when the d$_{xz}$/d$_{yz}$ degeneracy is maximally lifted \ldots{} while a null result \ldots{} would refute the mechanism.'' These are the Novelty-scout and mechanism-simulation functions the \swm{} externalizes, carried out here inside the backbone's own reasoning.

\paragraph{What the mid-capability backbone does instead.} qwen3.5-9b's reasoning is longer, at mean $13.0$k against $4.9$k characters, and broader rather than deeper. It enumerates $12$--$21$ candidate ideas, each with a one-line novelty tag such as ``a bit common,'' ``well-studied,'' or ``very popular, maybe less novel now,'' and rarely stress-tests the one it finally selects. The structured novelty-and-feasibility pass that \swm{} imposes is largely absent from its unaided trace, which is consistent with \swm{} supplying structure this backbone does not generate on its own.

\paragraph{Why we read the probe as illustrative.} Three limits prevent a stronger claim. (i)~Thinking was \emph{enabled} for this probe and \emph{disabled} when the \swm{} experiments and F21 ran (Appendix~\ref{app:repro}), so the probe shows that the frontier model \emph{can} reason this way, leaving open whether it did during those runs. (ii)~Both models perform some novelty self-assessment, and the difference lies in depth and structure rather than presence, which stops short of a clean strong-versus-weak dichotomy. (iii)~The sample is small, covering two backbones, eight subfields, and thinking-on re-runs. We therefore read the probe as \emph{consistent with} an internalization account rather than as evidence for it. A frontier backbone's unaided reasoning already resembles a structured novelty-and-mechanism check, which is compatible with an external \swm{} adding little on top, while a mid-capability backbone's more diffuse reasoning leaves room for the structure \swm{} imposes. A controlled test with matched thinking configuration, matched compute, and a placebo bad-draft probe is left to future work.

\end{document}